\documentclass{article}

\usepackage{lineno}

\usepackage[preprint]{neurips_2023}

\PassOptionsToPackage{numbers, compress}{natbib}

\usepackage[utf8]{inputenc} 
\usepackage[T1]{fontenc}    
\usepackage[colorlinks=true,
linkcolor=blue,
urlcolor=blue,
citecolor=blue]{hyperref}       
\usepackage{microtype,mathtools}      %

\usepackage{graphicx}

\usepackage{amsmath,amssymb,amsthm}
\usepackage{bbm}
\usepackage{float}
\usepackage{accents}
\usepackage{algorithm}

\usepackage{color,xcolor}

\usepackage{wrapfig}

\usepackage{wrapfig}
\usepackage{subcaption}
\usepackage{lipsum}
\newtheorem{thm}{\bf Theorem}
\newtheorem{claim}{\bf Claim}
\newtheorem{assumption}{\bf Assumption}

\newtheorem{rmk}{\bf Remark}
\newtheorem{lemma}{\bf Lemma}

\newcommand{\beqq}{\begin{equation}}
\newcommand{\eeqq}{\end{equation}}

\usepackage[utf8]{inputenc} 
\usepackage[T1]{fontenc}    
\usepackage{hyperref}       
\usepackage{url}            
\usepackage{booktabs}       
\usepackage{amsfonts}       
\usepackage{nicefrac}       
\usepackage{microtype}      
\usepackage{xcolor}  

\title{A Framework for Quantifying How Pre-Training and Context Benefit In-Context Learning}

\author{Bingqing Song\\
University of Minnesota\\
\And
Jiaxiang Li\\
University of Minnesota\\
\And
Rong Wang \\
University of Minnesota\\
\And
Songtao Lu\\
The Chinese University of Hong Kong\\
\And
Mingyi Hong\\
University of Minnesota
}

\nolinenumbers

\begin{document}

\maketitle


\begin{abstract}
Pre-trained large language models have demonstrated a strong ability to learn from context, known as in-context learning (ICL). Despite a surge of recent applications that leverage such capabilities, it is by no means clear, at least theoretically, how the ICL capabilities arise, and in particular, what is the precise role played by key factors such as pre-training procedure as well as context construction. In this work, we propose a new framework to analyze the ICL performance, for a class of realistic settings, which includes network architectures, data encoding, data generation, and prompt construction process. As a first step, we construct a simple example with a one-layer transformer, and show an interesting result, namely when the pre-train data distribution is different from the query task distribution, a properly constructed context can shift the output distribution towards the query task distribution, in a quantifiable manner, leading to accurate prediction on the query topic. We then extend the findings in the previous step to a more general case, and derive the precise relationship between ICL performance, context length and the KL divergence between pre-train and query task distribution. Finally, we provide experiments to validate our theoretical results.
\end{abstract}

\section{Introduction}\label{sec:intro}
Large language models \citep{devlin2018bert,radford2019language,brown2020language} are pre-trained on various texts to predict the next masked token. It is known that the pre-trained language models (LM)  possess strong in-context learning (ICL) capabilities. Specifically, in the inference stage, when the LM is provided with a sequential prompt, which consists of a few related examples and a query, the prediction accuracy can be significantly improved as compared to simply inputting a plain query. Such a kind of capability is intriguing, but so far it is by no means clear why it arises, and how to analyze it. 

Recently, there has been extensive research trying to understand and interpret the power of ICL through analyzing the structural property of LM, that is, how model structures such as attention mechanism in the Transformers can induce ICL capabilities
\citep{zhang2023trained,huang2023context,von2023transformers,dai2022can,olsson2022context,han2023explaining,ahn2024transformers,akyurek2022learning,yang2022transformers,mahankali2023one,li2023transformer,xing2024benefits}. Some other works show that Transformers can benefit ICL with the idea of  ``chain-of-thought'', by decomposing contexts into intermediate steps \citep{li2024dissecting,wei2022chain}. Some recent works quantify the {role of} pre-train task diversity {for ICL} when the pre-train distribution is different from query task distribution \citep{raventos2024pretraining}. 
Below let us discuss a few sets of representative theoretical works. 

The first line of works investigates the convergence and approximation power of Transformers in ICL \citep{zhang2023trained,huang2023context,chen2024training,kim2024transformers}. Transformer is utilized to approximate a linear regression model. However, instead of directly analyzing the ICL performance without changing any network parameters, some \textit{gradient-based} algorithms are typically manually implemented to optimize the loss function related to a prompt modeled by the Transformer. Further, the prompt is constructed by stacking on the embedding dimension of each query and answer; see Section {\ref{subsec:compare}}. 
for detailed discussions. 
Unfortunately, these settings do not represent practical use cases of ICL, where the prompt is constructed by stacking all the queries and answers {\it in a sequence} (see \eqref{eq:prompt}), and for a given context prompt, ICL is conducted {\it without} any parameter update. 

The second line of works focuses on characterizing the implicit implementation of algorithms when the prompt is fed into a single-layer Transformer. In \citet{von2023transformers}, it is shown that when the weight matrices in the Transformer have a certain structure and value, ICL with prompt implicitly performs one step of GD algorithm. In \citet{ahn2024transformers}, it is proved that the optimal parameters of a Transformer (single or multi-layers) can implement a step of preconditioned GD for ICL. \citet{edelman2022inductive,olsson2022context} claim that the attention mechanism implicitly learns information from inputs.  In \citet{han2023explaining}, ICL with Transformer structure is interpreted as a kernel regression problem. \citet{fu2023transformers} proves that Transformers learn to implement higher-order optimization methods to perform ICL. Although these works do not include the in-context pre-training, the prompt construction is the same as the previous line of works, therefore still drifting away from the real case.

Additionally, some other works explain ICL via Bayesian theory or Bayesian algorithm \citep{muller2021transformers,ahuja2023context,zhang2023and,wu2023many}. For example, \citet{xie2021explanation} studies how ICL benefits the prediction when the context examples in the prompt share the same concept with the query data. \\
\textbf{Our Contribution:}
As we have emphasized, there is still a lack of thorough theoretical understanding about how and why ICL works. Existing works that attempt to answer these questions are mostly conducted under either over-simplified or convenient but unrealistic settings {(as discussed above)}, making the results less relevant. In view of this, our work makes the following contributions to the ICL literature:

\vspace{-0.1cm}
\noindent{\bf(1)} A new framework is proposed, under which we provide analysis of the ICL performance. The  framework consists of specifications about network architectures, data encoding, generation, and prompt construction processes, which we believe is a set of more {\it realistic} settings comparing to what has been analyzed in existing works. 

\vspace{-0.1cm}
\noindent{\bf(2)} We build an example following our framework, theoretically and empirically demonstrating that context helps shift the pre-train distribution to query task distribution 
after passing through a trained Transformer, leading to higher accuracy in prediction.

\vspace{-0.1cm}
\noindent{\bf(3)} We consider the general case under our framework, and quantify the precise connection between ICL performance, context length, and KL-divergence between pre-train and query task distributions. Overall, our work provides a new, and more direct understanding of how pre-trained data distribution, and the construction of context influence the ICL performance.

\vspace{-0.2cm}
\section{The Proposed Framework of Modeling Data Generation and Prediction}\label{sec:framework}

\begin{figure}[h]
    \centering
    \includegraphics[ width=0.6\textwidth]{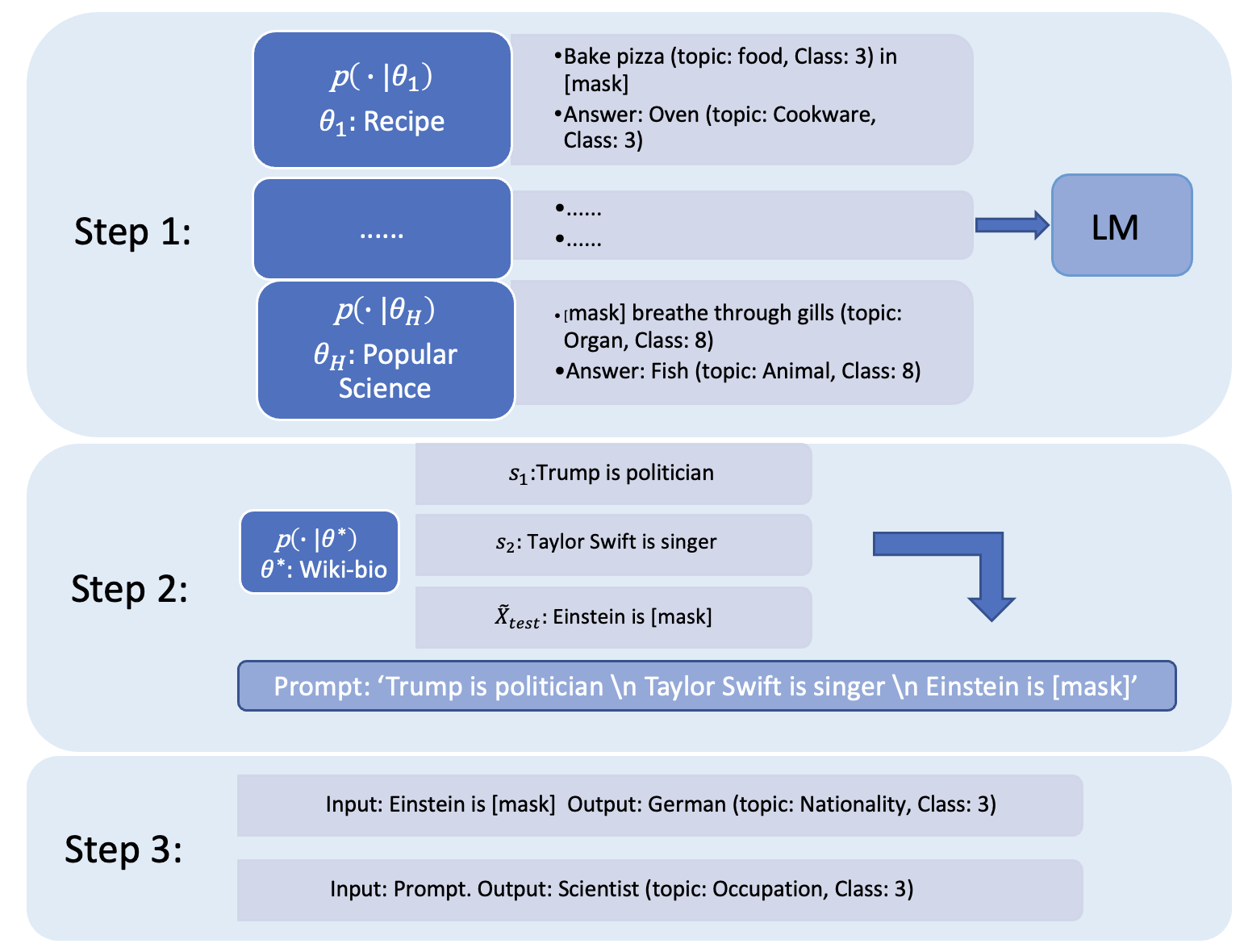}
    \caption{
    Summarization of the steps of ICL with latent concept generation, where each word consists of two attributes, i.e, \texttt{topic} and \texttt{class}. It 
    follows standard ICL procedure with $3$ steps: pre-training, prompt construction, and in-context inference. Following the setting in \citet{xie2021explanation}, we specialize the pre-training data and prompt generation process, so that they are conditioned on concepts. 
    Intuitively, the concept defines a distribution of generated sequence, which is specified in Section \ref{subsec: concept}.} 
    \label{fig:frame}
\end{figure}

{As discussed in Section \ref{sec:intro}, existing approaches to modeling ICL are often done under the settings that are a departure from the real ICL setting. In this section, we introduce a novel framework designed to approximate the realistic ICL process. Our proposed framework of ICL process modeling is summarized in Fig.~\ref{fig:frame}. In the following section, we will introduce our ICL modeling framework in detail. This framework comprises two components:  1) Modeling the language data generation process;} 2) Modeling the context construction prediction with pre-trained model. These two components are critical in modeling the ICL process, as they collectively allow us to analyze how changes in input—whether with or without context—affect the output distribution of the pre-trained model. Specifically: 1) It is essential to generate sequences that accurately represent the ground-truth data as context and query. This enables us to evaluate whether incorporating context samples during ICL leads to outputs that more closely resemble the ground truth; 2) The modeling of the prediction process is essential, as it determines how the pre-trained model utilizes context samples to generate responses. By analyzing the model’s behavior during inference, we can assess whether and how the inclusion of context influences the output distribution, ultimately leading to more accurate predictions that align with the ground-truth data.

.

\vspace{-0.5cm}


\subsection{Modeling data generation with latent concepts}\label{subsec: concept}
According to the discussion above, to accurately model the effect of ICL, it is important to model the generation of ground-truth data. A natural idea follows the latent concept generation in \citet{xie2021explanation}.
Instead of assuming that data is generated by a linear regression model, as has been done in a number of existing works \citep{huang2023context,ahn2024transformers}, the latent concept generation is more realistic since it more accurately characterizes the  correlation between tokens and provides a concrete way of expressing the distributions to be predicted. Such setting has been widely adopted in topic modeling and NLP analysis \citep{blei2003latent,gruber2007hidden}. 

{\bf Latent concepts.}
Following the setting in \citet{xie2021explanation,wang2023ICL}, we assume that each sequence (in pre-training or inference) is generated by a latent concept $\theta\in{\mathbf{\Theta}}$, where $\mathbf{\Theta}$ is a family of concepts. For example, the latent concept can be explicitly defined as `wiki-bio,' indicating that the associated sequence represents biographical information extracted from Wikipedia. Each concept $\theta$ has an associated sequence generation distribution $p(\cdot|\theta)$, which defines the probability distribution over generated sequences conditioned on $\theta$.

\begin{figure}[h]
   \centering
    \includegraphics[clip, trim={0cm 7cm 13.1cm 0cm}, width=0.8\linewidth]{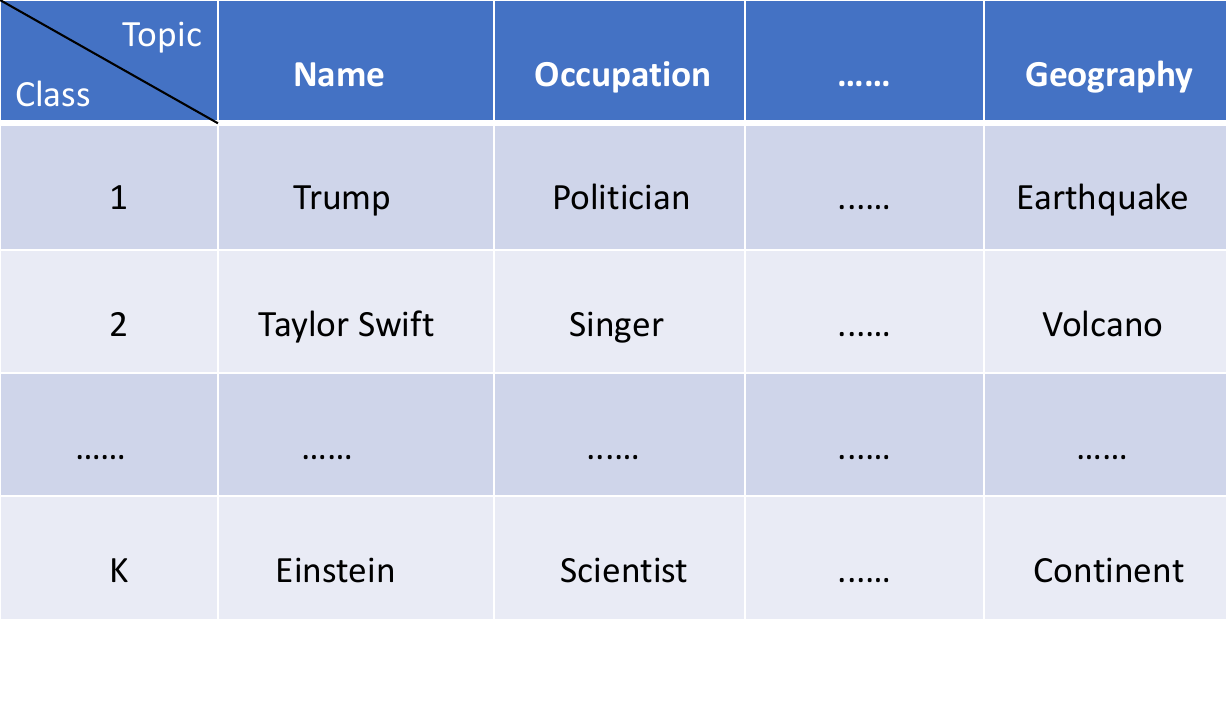}
     \vspace{-0.8cm}
    \caption{Each word has two attributes: topic and class. For simplicity, we assume each topic consists of $K$ different classes, each with only one word.} 
     \vspace{-0.4cm}
    \label{fig:token}
\end{figure}

{\bf Key token attributes.}
We assume the generated sample is a sequence of key words, and leave out the tokens that do not contain specific meaning. As an example, a generated sequence `Trump is a politician.' is reduced to `Trump politician'. Further, we assume each token consists of two attributes: \texttt{topic} and \texttt{class}. 
The idea of involving two attributes originates from the tabular data formulation \citep{fang2024large}, where each token is accurately represented by its corresponding attributes in the table; see Fig. \ref{fig:token} for an illustration. 
The topic attribute is straightforward and intuitive. On the other hand, the class attribute assigns a numerical label to each token within a given topic. If two topics are related, we assume that their corresponding tokens share a one-to-one mapping, meaning that each token in one topic has a direct counterpart in the other. This correspondence is encoded through the same class number. For example, `Name' and `Occupation' are related topics. We know the occupation of Trump (Class 1) is politician, then the class number of `Politician' has to be 1, which is the same as `Trump' (see Fig. \ref{fig:token}). This representation offers a straightforward means to quantify the distribution of samples across different attributes, enabling better analysis and understanding of the data structure. 
 For simplicity, throughout the paper, we assume that each word belongs to exactly one topic and class, but this requirement can be relaxed. 

\begin{rmk}
We restrict the output to key tokens and assign structural attributes to each key token. This approach serves as an approximation of real-world sequence generation by capturing the core meaning of the content, much like how tabular data distills essential features for analysis. This method, which has been widely explored in machine learning \citep{hegselmann2023tabllm}, enables efficient processing of critical information. 
\end{rmk}
\vspace{-0.3cm}



{{\bf Example: Sequence generation based on latent concepts and token attributes.}
With the latent concept model and token attributes in place, let us consider how language sequences are generated. When the sequence generation is about the biography of a person, the current concept is $\theta$=`wiki-bio', and
the generation process of sequence $s$ can be summarized in two steps: \\
{\bf Step 1 (First token)}: Given the concept `wiki-bio', the first token is likely about a topic highly related to the concept, e.g, name. However, it is totally random which name it will generate.  Suppose the first generated token is `Trump'. This indicates that the first token is likely to be with \texttt{topic} attribute related to $\theta$, while the \texttt{class} attribute is random.\\
{\bf Step 2 (Subsequent tokens)}: Suppose the first token is `Trump', the subsequent tokens are likely to remain within the `wiki-bio' concept, covering relevant topics such as occupation, birth year, or notable achievements. However, within each topic, the specific tokens generated are highly influenced by their association with first token `Trump'. For instance, within the occupation topic, `politician' is significantly more probable than `singer', as it aligns with the factual attributes of the entity. The above procedure implies the following tokens are likely to be with \texttt{topic} attribute related to $\theta$, while the \texttt{class} attribute tends to be consistent with the first token.\\
{\bf Modeling real-world generation.} Inspired by the example above, we consider the generation process of a length $N$ sequence $\boldsymbol{w}:=w_{1:N}$, where $w_i,i\in[N]$ denotes the $i$-th toke in $\boldsymbol{w}$.  The sequence generation distribution $p(\cdot|\theta)$, which involves both topic and class distribution, can be modeled as following:\\
{\bf Step 1 (First token $w_1$)}: The \texttt{topic} attribute $t$ is sampled from a topic distribution conditioned on the given concept, expressed as $t\sim p_t(\cdot|\theta)$, where $p_t$ denotes the topic distribution. The \texttt{class} attribute $k$ is generated on a random distribution without being conditioned on $\theta$, expressed as $k\sim p_c$, where $p_c$ denotes the class distribution.\\
{\bf Step 2 (Subsequent tokens $w_{2:N}$)}: 
Similar to {Step 1}, the \texttt{topic} attribute $t$ follows the distribution conditioned on $\theta$, i.e, $t\sim p_t(\cdot|\theta)$. The \texttt{class} attribute generation is conditioned on the class of first token. Suppose the class of first token is $k^*$, we assume the class $k$ of following tokens is generated by the following distribution:
\begin{align}\label{eq:classdef}
p_c(k^*\mid w_1)=Q,\; p_c(k\mid w_1)=\frac{1-Q}{K-1},k\neq k^*,
\end{align}
where $K$ is the total number of classes, and $Q\in[0,1]$ is close to $1$, so that $k^*$ is more likely to occur than the rest of classes. $p_t$ and $p_c$ together induce $p(\cdot|\theta)$.

We justify the above modeling method in Remark \ref{rmk:data} in Appendix \ref{sec:gene}.

{\bf Sequence Generation.} With the above modeling framework of data generation, we can formulate the generation of pre-train, context and query sequences. We assume there are $H$ different pre-train datasets $R_h,h\in [H]$, where each of the data is generated in the aforementioned way, i.e. each with $n_1$ sequences generated by latent concept $\theta_h$, expressed as 
$R_{h,i}\sim p\left(\cdot\mid \theta_h\right),\;i\in[n_1].$
Besides, in the ICL setting, we require context sequences to establish the context for the model. 
We assume  there are $n$ context samples generated by the concept $\theta^*$, i.e, $s_i\sim p\left(\cdot\mid \theta^{*}\right),\;i\in[n].$
Similarly, the query $s_{\text q}$ is also generated as a sequence of words by the distribution $p(\cdot\mid\theta^*)$, while
the last (few) tokens in $s_{\text{q}}$ are manually masked out after the generation, which become placeholder to predict. Denote the masked $s_{\text{q}}$ as  
$\widetilde{X}_{\text{q}}:=\left(X_{\text{q}},\textrm{[mask]}\right)$, while the original query sequence can be written as  $s_{\text q}=(X_{\text q},y_{\text q})$. As an example, suppose we have the query sequence $s_{\text q}=\text{`Trump politician American'}$, and we mask out the last word `American', which becomes a placeholder to predict. $s_{\text q}$ can be rewritten as $s_{\text q}=(\text{`Trump politician'},\text{`American'})$, where  $X_{\text q}=\text{`Trump politician'}$ and $y_{\text q}=\text{`American'}$. The masked $s_{\text q}$ is denoted as  $\widetilde{X}_{\text{q}}:=\left(\text{`Trump politician'},\textrm{[mask]}\right)$. Similarly, we can split any context sample $s_i$ into two parts $X_i$ and $y_i$ to align with the formula of $s_{\text q}=(X_{\text q},y_{\text q})$. We assume the sequence lengths of the pre-trained data and context samples of $R_{h, i},s_i, s_{\text {q}}$ and $\widetilde{X}_{\text {q }}$ are all $N$, {which is generic since we can always achieve this by truncating and padding.}


\subsection{Modeling ICL prediction with pre-trained LM}\label{ssubsec:pretrain}
With the above modeling of data generation, the next task is to model how pre-trained model generates the output. First, let us consider the case without ICL.\\
{\bf Masked output prediction} Given an LM $M$ pre-trained on $H$ different datasets $R_h,h\in [H]$ 
Given the masked query sequence $\widetilde{X}_{\text q}$,
the task of LM (with or without in-context learning) is to predict the last few masked words in $\widetilde{X}_{\text 1}$, which is denoted as $y_{\text q}\in\mathcal{Y}$, and $\mathcal{Y}$ is the output domain. The prediction probability based on the model $M$ is naturally defined  as
$
P_M (y\mid \widetilde{X}_{\text q})
$, which can be understood as predicting the masked output $y$ based on model $M$ and prompt $X_\text{query}$.
Suppose a language model $M$ completely learns the pre-train tasks, we can further simplify the expression of prediction probability as
$P(y\mid R_{1:H}, \widetilde{X}_{\text q}).$
\begin{rmk}
LM can face several key challenges in generating accurate and contextually appropriate sequences: First, in the absence of context, the model may struggle to determine the appropriate topic for generated tokens. Second, without context samples, the model generates output solely on its pre-trained distribution rather than adapting to task-specific distribution. Thus, the context provided to LM is critical to prediction. In the following, we introduce the process of making predictions using LM with ICL and discuss how to properly construct context.
\end{rmk}
{\bf Masked output prediction with ICL.}
Recall that when LM fully learns pre-train tasks $R_{1:H}$, the expression of prediction probability is $P(y\mid R_{1:H}, \widetilde{X}_{\text{q}})$. Following this, suppose another $n$ context samples $s_{1:n}$ are provided, the prediction is conditioned on both  $\widetilde{X}_{\text{q}}$ and $s_{1:n}$, which is denoted as $P(y\mid R_{1:H}, s_{1:n},\widetilde{X}_{\text{q}})$. The prediction is denoted as $\widehat{y}_{\text q}$,  and the explicit expression of $\widehat{y}_{\text q}$ depends on the problem, e.g, regression or discrete token prediction.}


{{\bf Context construction.} Recall that each context sample $s_i$ can be decomposed as  $X_i$ and $y_i$, i.e., $s_i = (X_i,y_i)$.  
We define the {\it stacked prompt}, denoted as $\mathbf{Z}$
 as:
\begin{align}\label{eq:prompt}
\mathbf{Z}_{\rm stacked}&:=(s_{1:n},\widetilde{X}_{\text q})=\left(X_1,y_1,\cdots,X_n,y_n,X_{\text q},\textrm{[mask]}\right).
\end{align}
The constructed prompt can be directly fed into an LM to make output prediction.
\begin{rmk}
In Section \ref{subsec: concept} and \ref{ssubsec:pretrain}, we have introduced two modeling methods: First, we model the natural language generation by latent concept model with topic and class attributes; second, we model the token prediction using LM with ICL. The modeling framework can be summarized in Fig.~\ref{fig:frame}. With this framework, we can explicitly characterize how context improves LM prediction by analyzing its impact on sequence generation. Specifically, our framework allows us to quantify the role of context in LM predictions through the following key aspects: 1) The sequence distribution $p(\cdot|\theta)$ can be defined in closed-form; 2) The constructed context is formally represented in a closed-form expression. This allows us to quantify the influence of context on LM predictions and understand which aspects of context are most crucial for enhancing prediction quality.  
\end{rmk}}
\vspace{-0.3cm}
{\subsection{Comparing proposed context construction with existing works}\label{subsec:compare}
\vspace{-0.2cm}
We emphasize that our context construction method is different, and more realistic, as compared with existing approaches. We argue that the prompt construction in \eqref{eq:prompt} is natural and consistent with how ICL context is constructed in practice; for example, we refer the readers to the implementation of  the well-known MetaICL framework \citep{min-etal-2022-metaicl}\footnote{Code available at: \url{https://github.com/facebookresearch/MetaICL}}, where the exact construction has been used. To illustrate this, in this section, we provide a simple example demonstrating how our structured context formulation method better align with practice.  

For simplicity, we assume the token embedding dimension as $D$;  $X_i$ and $y_i$ with length $N$ and $1$, respectively. Then we have $
\mathbf{Z}_{\rm stacked}\in\mathbb{R}^{D\times \left((n+1)\cdot N\right)}$. 
Different from \eqref{eq:prompt}, which does not make any assumptions between the input $X_i$'s and output $y_i$'s, in the existing literature which analyzes the ICL performance \citep{huang2023context,ahn2024transformers}, the ground-truth model is assumed to be {\it linear}: $y_i=WX_i,X_i\in \mathbb{R}^{D\times1}, W\in\mathbb{R}^{1\times D}$, where $y_i$ is scalar. 
\begin{align}\label{eq:fakeprompt}
\mathbf{Z}_{\rm linear}=\left(s_{1:n},X_{\text q}\right)=\left(\begin{array}{ccccc}
X_1 &  \cdots & X_n & X_{\text q} \\
y_1 & \cdots & y_n & \mathbf{0}
\end{array}\right)
\end{align}
In next few paragraphs, we will soon explain that the two prompt constructions \eqref{eq:prompt} and \eqref{eq:fakeprompt} make huge difference in terms of how the context impacts the prediction.  

\noindent\textbf{Transformer Structure:} Let us consider a standard one-layer transformer without embedding layer for analysis. 
For simplicity, we do not consider skip connection and normalization. The simplified setting is standard in literature \citep{zhang2023trained,huang2023context,li2023transformers}. Let us denote the input as $\mathbf{Z}\in \mathbb{R}^{L\times G}$ , where $L$ and $G$ are appropriate dimensions depending on how input is structured. Suppose the prediction is the output of the following Transformer:
\begin{align}\label{eq:transformer}
f_{\bf w}(\boldsymbol{Z})=(\boldsymbol{W}^V \boldsymbol{Z}) \sigma\left({(\boldsymbol{W}^K \boldsymbol{Z})^{\top}(\boldsymbol{W}^Q \boldsymbol{Z})}/{\sqrt{L}}\right),
\end{align}
where ${\bf w}:=\{\boldsymbol{W}^Q,\boldsymbol{W}^K,\boldsymbol{W}^V\in\mathbb{R}^{L\times L}\}$ are query, key, value matrix, respectively. The activation function $\sigma(\cdot):\mathbb{R}^{G\times G}\mapsto(0,1)^{G \times G}$ is a column-wise softmax activation. Define $A(\mathbf{Z})$ as the attention kernel, i.e, $A(\mathbf{Z}):=\sigma\left((\boldsymbol{W}^K \boldsymbol{Z})^{\top}(\boldsymbol{W}^Q \boldsymbol{Z}) / \sqrt{L}\in\mathbb{R}^{G\times G}\right)$.

\textbf{Comparison of masked output prediction with prompt \eqref{eq:prompt} and \eqref{eq:fakeprompt}}: 
In the following example, we compare the explicit output prediction with prompts in \eqref{eq:prompt} and \eqref{eq:fakeprompt} and the Transformer structure in \eqref{eq:transformer}. To make the comparison more clear and intuitive, consider the special case with following assumptions:
\begin{assumption}\label{ass:example}
(i) The embedding for the masked token in \eqref{eq:prompt} is $\mathbf{0}\in\mathbb{R}^{L\times 1}$; (ii) The attention kernel is uniform, i.e, $A(\mathbf{Z})_{ij}=\frac{1}{G}$; (iii) Sequence length $N=2$.
\end{assumption}

With Assumption \ref{ass:example}, we show the two predictions based on different prompt constructions as follows.

\noindent\textbf{(a) Prediction from \eqref{eq:prompt}:} Set $\mathbf{Z}=\mathbf{Z}_{\text{stacked}}$ in \eqref{eq:transformer}. In this case, $L=D,G=2n+2$. We have 
\begin{align}
&\widehat{y}_{\text {q}}=f_{\bf w}(\mathbf{Z}_{\text{stacked}})_{1:D, 2n+2}=
\frac{1}{2n+2}\sum\limits_{i=1}^{n}\boldsymbol{W}^V(X_i+y_i)+\frac{1}{2n+2}\boldsymbol{W}^V{X}_{\text {q}}.\label{eq:truepred}
\end{align}
where $f_{\bf w}(\cdot)_{a:b,c:d}$ denotes the submatrix of $f_{\bf w}(\cdot)$ with $a$-th to $b$-th row, and $c$-th column to $d$-th column.\\ 
\noindent\textbf{(b) Prediction from \eqref{eq:fakeprompt}:} Set $\mathbf{Z}=\mathbf{Z}_{\text{linear}}$ in \eqref{eq:transformer}. In this case, $L=2D,G=n+1$. We have
\begin{align}
\widehat{y}_{\text q}=f_{\bf w}(\mathbf{Z}_{\text{linear}})_{D+1:2D,n+1}=\sum\limits_{i=1}^{n}\frac{1}{n+1}y_i.\label{eq:fakepred}
\end{align}
\begin{rmk}
Notably, the above two predictions coming from two ways of prompting construction have completely different meanings. In \eqref{eq:truepred}, the prediction $\widehat{y}_{\text {q}}$ is a weighted sum of the $\boldsymbol{W}^VX_i$,$\boldsymbol{W}^Vy_i$ and $\boldsymbol{W}^VX_{\text {q}}^{\top}$, which include information from both in-context inputs and outputs. However, in \eqref{eq:fakepred}, the output is a weighted sum of the in-context outputs $y_i$ only, while the weights are determined by the correlation between $X_i$ and $X_{\text q}$. Additionally, we justify Assumption \ref{ass:example} in Remark \ref{rmk:promptcompare} in Appendix \ref{sec:gene}. 
\end{rmk}

}

\section{A Statistical Model: Pre-trained Transformer Shifts the Output Distribution }\label{sec:example}
\begin{figure}[h]
	\begin{center}
        \includegraphics[width=0.8\linewidth]{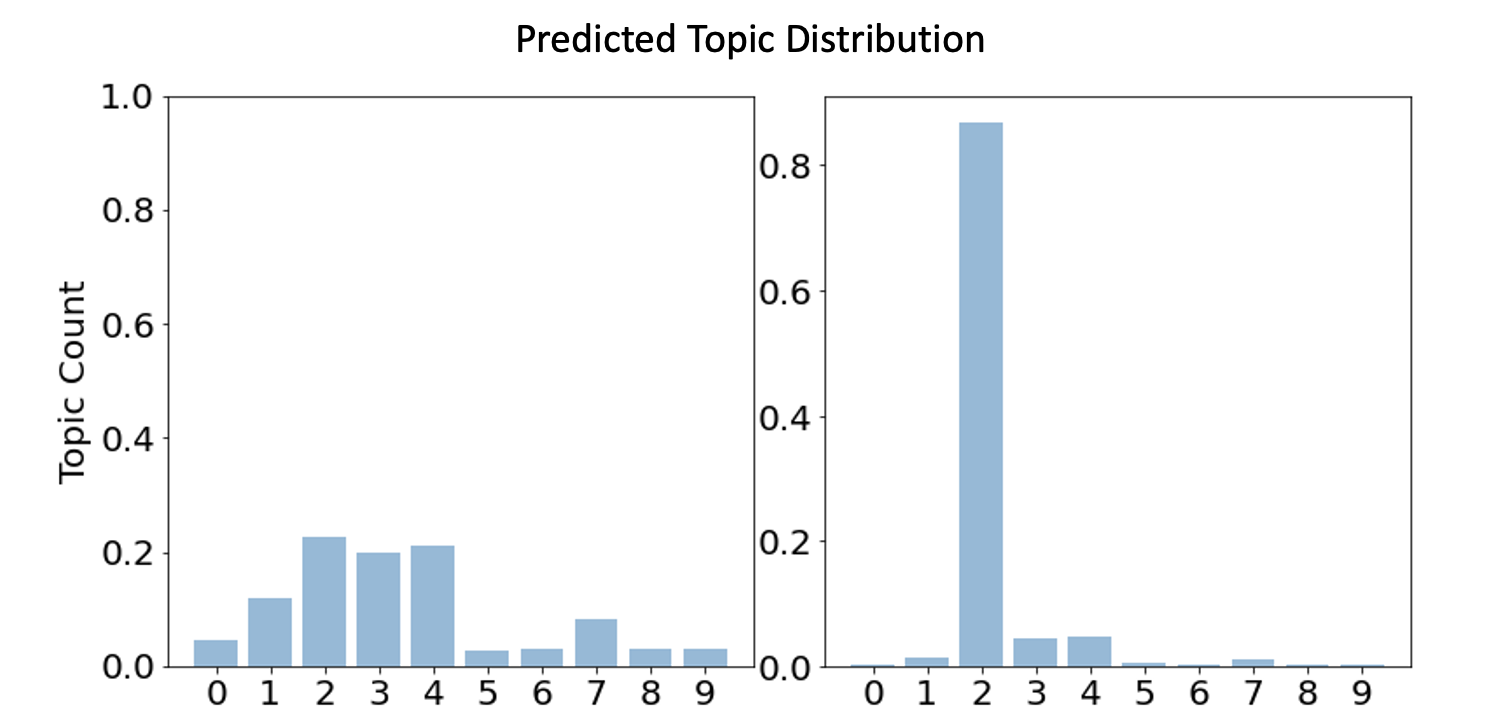}
	\end{center}
	\caption{The distribution of topics in LDA example. Left: Prediction without ICL. Right: Prediction with ICL. Given $10$ different topics, denoted as `0' to `9'. The topic of the query task is `2'. When the prediction is made by $f_{\bf w}(\widetilde{X}_{\text q})$, the topic distribution does not lean toward `2'. However, when the prediction is made by $f_{\bf w}(\textbf{Z}_{\text{stacked}})$, it is more likely that the query topic `2' is predicted. }
	\label{fig:dist}
\end{figure}

So far we have proposed a framework to quantify the effect of context in ICL. In this section, we construct a concrete example (referred to as a ``modified LDA (Latent Dirichlet Allocation)'' setting), to show that a trained Transformer can learn context samples, shift the output distribution to the query task distribution, and achieve higher accuracy in prediction than the case without ICL. In the following, we will demonstrate the example under the setting in Section \ref{sec:framework}. To intuitively understand our example, we first show our synthetic result based on our example in Fig.~\ref{fig:dist}, in which we display the distribution of topics in the prediction (with or without ICL) when the target topic is `2'. As indicated in the figure, ICL greatly improves the accuracy of the topic in prediction. 
Our detailed experiment setting is available in Appendix \ref{sec:synthetic}.  
\subsection{Data Generation and Encoding}\label{sec:data}
We consider a specific data generation procedure following Section \ref{subsec: concept}, which is a modified version of the Latent Dirichlet Allocation model (LDA) \citep{blei2003latent}. The standard LDA setting is commonly used in topic modeling of NLP tasks \citep{jelodar2019latent,li2023transformers}. In our modified LDA setting, the ground-truth vocabulary consists of $T$ topics, each with $K$ different classes, and each class contains only one word. Each sample $\boldsymbol{w}:=w_{1:N}$ (training,context and query) is a sequence of $N$ tokens, which is generated by the following procedure: (a) Randomly choose $\tau$ different topics $t_1,\cdots,t_{\tau}$ from $T$ total topics. Generate concept $\theta\in\mathbf{\Theta}$; (b) Follows \textbf{Step 1} in Section \ref{subsec: concept} Part {\bf Modeling real-world generation}; (c) Follows \textbf{Step 2} in Section \ref{subsec: concept} Part {\bf Modeling real-world generation}.

We use the above procedure to generate the training, query and context sequences, while we specify the distribution $p_t$ and $p_c$ in the following.\\
\textbf{Generated Data Distribution: }For training, query and context sequences, the generated data distribution are different, to approximate the real-world case. Let $t(\cdot)$ and $c(\cdot)$ denote the topics and classes attributes of a sequence, respectively. The topic distribution of training sequence $\boldsymbol{w}$ given a $\theta$: 
\begin{align*}
p_t\left(t({w}_j)=t\right)=1/\tau,\; t\in\{t_1,\cdots,t_{\tau}\},\;\forall j\in[N],
\end{align*}
which implies the topic is uniformly distributed in corpus. For query and context sequences $\boldsymbol{w}$, the topic distribution is 
\begin{align}\label{eq:queryclass}
\begin{cases}
&P\left(t\left(w_j\right)=t\right)=1 / \tau,\;  t \in\left\{t_1, \cdots, t_\tau\right\}, j \in[L_1] \text {,}\\
&t\left(w_j\right)=t^{*}, j=L_1+1,\cdots,N.
\end{cases}
\end{align}
For all types of sequences, the class distribution of first token is $p_c\left(c(w_1)=t\right)=1/K$, while the class distribution of following tokens follows \eqref{eq:classdef}.

\textbf{Masking: } For training sequences, the masking probability of each token is $p_m$. For an original training sequence, randomly choose masked indices $\pi(\boldsymbol{w})\subset [N]$. The training is done by masked token prediction . For query sequence, mask the $L_2$ last tokens, while leaving the first $L_1$ tokens unmasked ($L_1+L_2=N$). The query task is to predict the last $L_2$ masked tokens. Recall query sequence $s_{\text q}=(X_{\text q}, y_{\text q})$, $X_\text{q}$ has length $L_1$ and $y_{\text q}$ has length $L_2$. The context samples are not masked. 
\vspace{-0.1cm}
\begin{rmk}
The data generation in \eqref{eq:queryclass} is an example of the query data generation in ICL framework defined in Section \ref{sec:framework}. In the inference phase, with the query word `Einstein', $p(\cdot\mid \theta_{\text q})$ is supposed to attend to the key topic `Occupation' in prediction. However, the predictor does not know the current prediction should attend to `Occupation'.
\end{rmk}
\textbf{`Two-hot' Encoding: } For any generated sequence $\boldsymbol{w}$ with length $N$, we define two associated matrices $U,\widetilde{U}\in\mathbb{R}^{(T+K+2)\times N}$ to denote the \textbf{encoded } binary matrix of sequence and masked sequence, respectively. The $N$ columns represent $N$ tokens in the sequence, and the rows are the indicators of mask, topics and classes, 
More specifically, each entry in $U$ (same for $\widetilde{U}$) can be written as following (See Fig.~\ref{fig:tableexample}) in Appendix \ref{sec:gene}:
\begin{align*}
\begin{cases}
&
U_{l,j}=1 \text { if } w_j=\textrm{[mask]}, \;l=0,\\
&U_{l,j}=1 \text { if } t(w_j)=l, \; 1\leqslant l\leqslant T,\\
&U_{l,j}=1 \text { if } w_j=\textrm{[mask]}, \;l=T+1,\\
&U_{l,j}=1 \text { if } c(w_j)=l-T-1,\; T+2\leqslant l\leqslant T+K+1,
\end{cases}
\end{align*}
while the other entries are $0$s. For any generated training sequence $s_{\text {train}}$, let $U_{\text{train}},\widetilde{U}_{\text{train}}$ denote the associated encoded binary matrices (unmasked and masked) with distribution $\mathcal{D}_{\text{train}}$ and $\widetilde{\mathcal{D}}_{\text{train}}$, respectively.  Recall the notation from Section \ref{subsec: concept}, for a query sequence $s_{\text q}=(X_{\text q},y_{\text q})$ and its masked version $\widetilde{X}_{\text q}$, we use $U_{\text q}$ and $\widetilde{U}_{\text q}$ to denote the encoded binary matrices (unmasked and masked), with distribution $\mathcal{D}_{\text q}$ and $\widetilde{\mathcal{D}}_{\text q}$, respectively.  And we denote the class of the first token in query sequence as $k^{*}_{\text q}$.
\vspace{-0.3cm}
\subsection{Training Objective and Inference}
The generated and encoded data are used to train a Transformer model by masked token prediction. In the following we specify the training and inference steps. Let $U_{:j}$ denote the $j$-th column of matrix $U$. We consider the same regularized $\ell_2$ loss function as in \cite{li2023transformers}:
\begin{align}\label{eq:obj}
L({\bf w})=\mathbb{E}_{{U} \sim \mathcal{D}_{\text{train}}} \mathbb{E}_{\pi} \frac{1}{|\pi|} \sum_{j \in \pi(U)} l\left(f_{\bf w}(\widetilde{{U}})_{: j}, {U}_{: j}\right)+\lambda\|{\bf w}\|_2^2,  
\end{align}
where $f_{\bf w}(\cdot)$ is defined in \eqref{eq:transformer},  $l(f_{\bf w}(\widetilde{U})_{: j}, U_{: j})=\|f_{\bf w}(\widetilde{U})_{: j}-U_{: j}\|^2$. The regularization term is applied due to the effectiveness of weight decay in training transformers, where $\lambda$ is the regularization weight.
Then we consider following two different types of prediction modeled by the Transformer structure in \eqref{eq:transformer}. Our goal is to predict the last $L_2=p_m\cdot N$ masked words in $\widetilde{U}_{\text q}$. We provide the explicit formula of the prediction of masked token with model $f_{\bf w}(\cdot)$ in Appendix \ref{sec:proofclaim}.
\vspace{-0.2cm}
\subsection{Pre-trained One-layer Transformer}
 We consider the one-layer 
 Transformer in \eqref{eq:transformer} pre-trained on $H$ different tasks. We consider a special case where the attention score depends on position only. To be specific, we have the following assumption:
\begin{assumption}\label{ass:attention}
The Transformer in \eqref{eq:transformer} has the following form of attentions:

(1) For an encoded matrix $U\in \mathbb{R}^{(T+K+1)\times N}$ (same for $\widetilde{U}$)
\begin{align*}
A({U})_{j',j}=1/N,\;j'\in[N], L_1+1\leqslant j\leqslant N.
\end{align*}
(2) For input ${\bf Z}_{\text{stacked}}=(s_{1:n},\widetilde{X}_{\text q})\in\mathbb{R}^{D\times N(n+1)}$, the associated encoded matrix ${U}_{\text{stacked}}$ has the following property:
\begin{align*}
&A\left({U}_{\text{stacked}}\right)_{(i-1)N+r,j}= a_i/N,\;i\in[n+1],r\in[N],
\end{align*}
where $L_1+1\leqslant j\leqslant N,a_1<\cdots<a_{n+1}, \sum\limits_{i=1}^{n}a_n=1$.
\end{assumption}
\vspace{-0.1cm}
In addition, we make the following generic assumptions on the document length and parameter $\boldsymbol{W}^V$.
\begin{assumption}
Assume the document length $N$ is infinity. 
\end{assumption}
\begin{assumption}
Assume $\boldsymbol{W}^V$ has block diagonal structure:
$\boldsymbol{W}^{V}=\left(\begin{array}{cc}
W_1 & \mathbf{0} \\
\mathbf{0} & W_2
\end{array}\right)$. 
\end{assumption}
\begin{rmk}
We assume the $\boldsymbol{W}^V$ to be block diagonal in order to keep the topic and class predictor independent. Similar block diagonal parameter assumption is also used in \citep{zhang2023trained,huang2023context}.
\end{rmk}
Let $\text{topic}(\cdot)$ and $\text{class}(\cdot)$ denote functions that select the most probable topic and  class, respectively, from a column vector of probability distributions over topics and classes. Given a column vector $b_t$
representing the probability distribution over topics and a column vector $b_c$ representing the probability distribution over classes, we define:
\begin{align*} 
\operatorname{topic}\left(b_t\right)=\arg \max _t p_t(t),\;\operatorname{class}\left(b_c\right)=\arg \max _c p_c(c)\end{align*}
\begin{claim}\label{thm:lda}
Generate each encoded pre-train data $U_{\text{train}}\sim\mathcal{D}_{\text{train}},\widetilde{U}_{\text{train}}\sim\widetilde{\mathcal{D}}_{\text{train}}$. There exists $a_1<a_2<\cdots<a_{n+1}$, such that if we train a one-layer Transformer \eqref{eq:transformer}: (1) With attention $A(\cdot)$ that satisfies Assumption \ref{ass:attention}; (2) By minimizing the $\ell_2$-regularized objective function \eqref{eq:obj} with variable $\boldsymbol{W}^{V}$. Suppose $\boldsymbol{W}^{{V}*}\in \underset{\lambda\rightarrow 0}{\operatorname{lim}}\operatorname{argmax}L(\boldsymbol{W}^V)$, and we use $f_{\boldsymbol{W}^{\boldsymbol{V}*}}(\cdot)$ as prediction model. Given an encoded masked query sequence $\widetilde{U}_\text{q}$, predict the last $L_2$ tokens ($L_1+1\leqslant j\leqslant N$) as $\widehat{y}_q$. If the input is $\widetilde{U}_{\text q}$ without context,
\vspace{-0.1cm}
\begin{align*}
P({\operatorname{topic}(\widehat{y}_{\text q})}_j=t^*)=1/T,\;\operatorname{class}(\widehat{y}_{\text q})_j=T+k^{*}_{\text q}+1;
\end{align*}
if the input is ${U}_{\text{stacked}}$,
\begin{align*}
\operatorname{topic}(\widehat{y}_{\text q})_j=t^{*},
\operatorname{class}(\widehat{y}_{\text q})_j=T+k^{*}_{\text q}+1.
\end{align*}

\end{claim}
\begin{rmk}
Claim \ref{thm:lda} shows a case where the pre-train data distribution is different from the query data distribution, where topic is uniformly distributed in the pre-train stage. While in the inference stage, the aim is to focus on topic $t^{*}$. Intuitively, if the prompt can provide context samples that focus on topic $t^{*}$, then the pre-trained model will tend to predict the word from topic $t^{*}$. This is because more recent sequences are assigned higher weights, making the class of the query sequence $s_{\text q}$ more influential than that of the context samples. Additionally, we justify Assumption \ref{ass:attention} in Appendix \ref{sec:gene} and provide proof in Appendix \ref{sec:proofclaim}.
\end{rmk}
\vspace{-0.5cm}
\section{Generalized Case: Quantify the effect of Context in ICL}
\vspace{-0.3cm}
In the Claim \ref{thm:lda}, we used a specific data generation following our proposed framework in Section \ref{sec:framework}, and theoretically prove that ICL can improve the prediction accuracy when the pre-trained model on corpus fails to capture the query topic distribution.  However, we still need to understand how trained Transformers and context improve prediction in a more general case. Here we ask the following question: In general, how to quantify the connection between the pre-training (including distribution, number of tasks, and number of samples) and the performance of ICL? In this section, under our setting in Section \ref{sec:framework}, we use a Bayesian framework following \citet{xie2021explanation} to quantify the relationship between the ICL prediction and prompt length, number of pre-train samples, and KL-divergence between pre-train and query task distribution. Let us consider a discrete output domain $\mathcal{Y}$ with the following several assumptions.
\begin{assumption}\label{ass:y_dinstinct}(Distinguishability of Output)
For some constant $\epsilon$, the following relation holds $\forall y\in \mathcal{Y}$:
\vspace{-0.2cm}
\begin{align*}
&p\left(y_{\text{q}} \mid s_{1:n}, X_{\text {q}}, \theta^*\right)>\max\limits_{y \neq y_{\text q}} p\left(y \mid s_{1:n}, X_{\text {q}}, \theta^*\right)\quad +\epsilon / p\left(\theta^*\right).
\end{align*}
\end{assumption}
Assumption \ref{ass:y_dinstinct} requires that the optimal $y_{\text q}$ can be distinguished from the other answers in domain $y\in\mathcal{Y}$.
\begin{assumption}\label{ass:concept_distinct} The conditions on KL-divergence hold:
\vspace{-0.2cm}
\begin{align}
c_1:=&\frac{1}{H} \sum_{h=1}^H KL\left(p\left(\cdot \mid \theta_h\right) \| p\left(\cdot \mid \theta^*\right)\right)-\nonumber\\
\quad &KL\left(p\left(\cdot \mid \theta_h\right) \| p(\cdot \mid \theta)\right)<0,\;\forall \theta\in\mathbf{\Theta}.\label{eq:overalldivergence}\\
c_2:=&-K L\left(p\left(\cdot \mid \theta^*\right) \| p(\cdot \mid \theta)\right)<0, \;\forall \theta\in\mathbf{\Theta}.\label{eq:promptdivergence}
\end{align}
\end{assumption}
Equation \eqref{eq:overalldivergence} in Assumption \ref{ass:concept_distinct} requires that the query task is closest to all pre-train tasks on average. Further, both \eqref{eq:overalldivergence} and \eqref{eq:promptdivergence} require the data distribution conditioned on the concept be distinguished. Intuitively, one way to increase the distinguishability is to increase the sequence length $N$.  
\begin{assumption}\label{ass:bound_variance}
For each generated sequence $s$ in pre-train tasks $R_h,\;h\in [H]$ and prompt $s_{1:n}$, we can find a variance bound $\sigma^2$, such that
$\operatorname{var}\left(\log \frac{p\left(s \mid \theta\right)}{p\left(s \mid \theta^*\right)}\right)\leqslant \sigma^2,\forall \theta\in \mathbf{\Theta}.$
\end{assumption}
\vspace{-0.2cm}
Assumption \ref{ass:bound_variance} assumes the bounded variance log-likelihood ratio of a sequence conditioned on any other concept and query task concept $\theta^{*}$. Intuitively, when the sequence length $N$ is small, the variance $\sigma$ is small. Thus, together with the distinguishability condition in Assumption \ref{ass:concept_distinct}, there exists a trade-off on the sequence length $N$. 

\begin{thm}\label{thm:bayes}
Suppose Assumption \ref{ass:y_dinstinct}, \ref{ass:concept_distinct}, \ref{ass:bound_variance} hold, then with high probability, the following holds:
\vspace{-0.2cm}
\begin{align*}
\underset{y}{\arg \max }\ p\left(y \mid R_{1:H},s_{1:n}, X_{\text          
    q}\right) = \underset{y}{\arg \max }\ p\left(y \mid X_{\text q},\theta^{*}\right),
\end{align*} 
suppose $n_1, H,n$ satisfies: $n_1H>\frac{9\sigma^2}{c_1^2},n>\frac{9\sigma^2}{c_2^2},$ and
$-\left(n_1 Hc_1^{\prime}+n c_2^{\prime}\right)>\log \frac{1}{\epsilon},$
where $c_1'=c_1+\frac{3\sigma}{\sqrt{n_1 H}}$, $c_2'=c_2+\frac{3\sigma}{\sqrt{n}}$.
\end{thm}
\begin{rmk}
Theorem \ref{thm:bayes} shows that, an in-context predictor with the pre-trained model can approximate the optimal Bayesian predictor. Specifically, it is required that the overall distribution of pre-train data is close to prompt distribution (Assumption \ref{ass:concept_distinct}), so that the pre-trained model can generalize well on query data. Further, if data is more distinguishable, i.e., $c_1$ and $c_2$ are small, then fewer number of pre-train and prompt samples are required to approximate the optimal predictor. These finding align with intuition.
\end{rmk}
\section{Experiment}
\subsection{Experiment settings}\label{sec:setting}
\begin{figure}
    \centering
        \includegraphics[width=0.8\columnwidth]{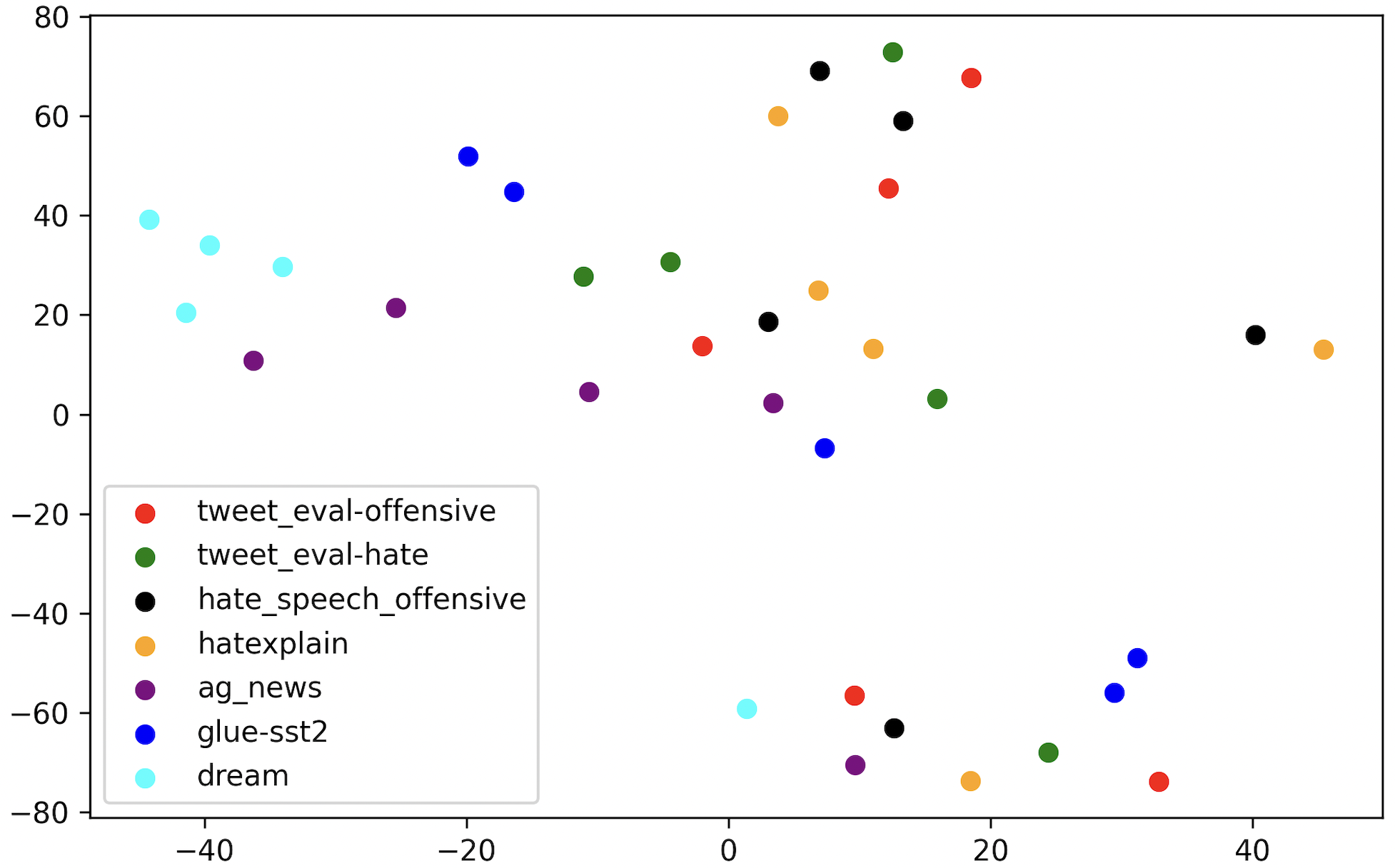}
	\caption{Distribution of concept token visualized by t-SNE plot. Intuitively, the first four tasks are similar. The visualization of the learned concept token indeed shows the representation of these four tasks are close each other compared to other tasks. 
    }
	\label{fig:concept_distance}
\end{figure}

In this section, we validate Theorem \ref{thm:bayes} with empirical results on GPT-2 models \citep{solaiman2019release} by showing that different pre-training tasks/datasets (that are more similar or dissimilar from the target task) can benefit the in-context inference, in terms of accuracy anf F1 score. Due to limited computation resources, we \textbf{do not pre-train} GPT-2 from scratch on various tasks to derive the pre-trained model $M$ in Section \ref{sec:framework}. Instead, we \textbf{fine-tune} the original GPT-2 with similar or dissimilar tasks to represent the final pre-trained model. We believe that this is already a strong showcase for Theorem \ref{thm:bayes} since we fine-tune from the same model but yield starkly different testing results after fine-tuning from the two different set of tasks.
Specifically, our experiment consists of three steps. \\
\textbf{Step 1:} Given $K$ available fine-tuning tasks/datasets, we first measure the similarity among them. We utilize Algorithm 1 in \citet{wang2023ICL}, where each tasks is assigned with `concept tokens' encoding the theme of the task. We then measure the distance between concept tokens to characterize the divergence between the tasks. The detail of how to obtain such `concept tokens' for each task is provided in Appendix \ref{sec:gpt}. 
In our experiment, we choose $K=7$ tasks: \texttt{hate\_speech\_offensive}, \texttt{hatexplain}, \texttt{tweet\_eval-hate}, \texttt{tweet\_eval-offensive}, \texttt{ag\_news}, \texttt{glue-sst2}, and \texttt{dream}. \\
\textbf{Step 2:} 
Once we obtain the concept tokens of each of the task, we simultaneously obtain a `representation vector' of each task by simply passing the concept tokens into the embedding layer of the GPT-2 model. We then could calculate the cosine similarity between these representation vectors. The obtained similarity scores 
are recorded in Table \ref{tab:concept_distance}. We also visualize the representation vectors of the concept tokens in Fig.~\ref{fig:concept_distance}. Intuitively, the following tasks \texttt{hate\_speech\_offensive}, \texttt{hatexplain}, \texttt{tweet\_eval-hate} and \texttt{tweet\_eval-offensive} are similar to each other, and they demonstrate similarity in Table \ref{tab:concept_distance}. We thus pick these four tasks as the fine-tune tasks which are similar to the target task to validate our Theorem \ref{thm:bayes}, whereas other datasets in the table would be consider as dissimilar tasks. \\ 
\textbf{Step 3:} Now we are ready to validate that similar tasks could boost the performance of the in-context inference target task. We fix the target task as \texttt{tweet\_eval-hate}, fine-tune the pre-trained GPT-2 by two set of other tasks, namely either with similar tasks (e.g. \texttt{hatexplain}) or dissimilar tasks (e.g. \texttt{glue-sst2}) from Step 2. Then we can evaluate the in-context inference accuracy and F1 score on these GPT-2 models fine-tuned on different tasks.

\vspace{-0.3cm}
\subsection{Results and Discussion}


\begin{table}[h]
\resizebox{\columnwidth}{!}{
\begin{tabular}{|l|l|l|l|l|l|l|l|l|}
\hline
Tasks & 1Diff & 1Sim & 2Diff & 1Sim1Diff & 2Diff1Sim & 1Diff2Sim & 2Diff2Sim & 1Diff3Sim \\ \hline
Macro-F1 & 56.4 & 59.1 & 57.1 & 60.2 & 59.0 & 63.1 & 59.9 & 60.7 \\ \hline
Accuracy & 0.59 & 0.62 & 0.62 & 0.63 & 0.60 & 0.64 & 0.62 & 0.63 \\ \hline
\end{tabular}}
\caption{Testing result of fine-tuning on different number of similar and dissimilar tasks. Here `aSimbDiff' means we fine-tune GPT-2 with a similar tasks and b dissimilar tasks.} 
\label{tab:Result}
\vspace{-0.1in}
\end{table}

\textbf{Main results:}
In our experiments, we pick up different number of similar and dissimilar tasks to fine-tune the GPT-2 model. What we expect, in the view of Theorem \ref{thm:bayes} is that with more similar fine-tune tasks (defined by the similarity score in Table \ref{tab:concept_distance}) 
we have better ICL testing performances. We report the performance over testing dataset of the target task of our fine-tuned models (fine-tune over different numbers of similar and dissimilar tasks) in Table \ref{tab:Result}. 

From Table \ref{tab:Result} we can conclude: (1) When only one dataset is used for fine-tuning, the model fine-tuned with similar tasks has 3\% higher accuracy than the model fine-tuned with dissimilar tasks; (2) In the case of using two datasets, the accuracy of the model using one similar and one dissimilar dataset for fine-tuning is 1\% higher than that of the model using two dissimilar datasets for fine-tuning, and the Macro-F1 score is 3.1 higher; (3) For the case of using three or four datasets for training, the performance of the fine-tuned model using more similar task datasets is better than that of the model using more dissimilar task datasets. These results indicate that fine-tuning the model with similar task datasets has a significant positive impact on the model.

In addition, we also conduct the same experiment using a different set of tasks (Table \ref{tab:concept_distance2}) and different model (GPT-XL), the resulting Table \ref{tab:Result2} consistently support our observation that using similar task datasets for fine-tuning has a significant positive impact on the model in terms of accuracy.

\begin{table}[h]
\centering
\resizebox{0.7\columnwidth}{!}{%
    \begin{tabular}{|l|l|l|l|l|l|}
        \hline
        Tasks & 2Sim & 2Diff & 4Diff & 4Diff1Sim & 4Diff2Sim \\ \hline
        yelp\_polarity & 91.1 & 87.8 &  88.9 &  90.7 &  91.2 \\ \hline
        imdb & 95.7 & 84.7 & 81.4 & 92.5 & 94.5\\ \hline
    \end{tabular}%
    }

\caption{Testing on two target datasets \texttt{yelp\_polarity} and \texttt{imdb} by fine-tuning GPT2-XL on different datasets, where the similarity is ranked according to the similarity score in Table \ref{tab:concept_distance2}. The reporting numbers are the accuracies on the testing datasets.} 
\label{tab:Result2}
\end{table}



\newpage

\bibliography{reference}

\begin{thebibliography}{37}
\providecommand{\natexlab}[1]{#1}
\providecommand{\url}[1]{\texttt{#1}}
\expandafter\ifx\csname urlstyle\endcsname\relax
  \providecommand{\doi}[1]{doi: #1}\else
  \providecommand{\doi}{doi: \begingroup \urlstyle{rm}\Url}\fi

\bibitem[Ahn et~al.(2024)Ahn, Cheng, Daneshmand, and Sra]{ahn2024transformers}
K.~Ahn, X.~Cheng, H.~Daneshmand, and S.~Sra.
\newblock Transformers learn to implement preconditioned gradient descent for in-context learning.
\newblock \emph{Advances in Neural Information Processing Systems}, 36, 2024.

\bibitem[Ahuja et~al.(2023)Ahuja, Panwar, and Goyal]{ahuja2023context}
K.~Ahuja, M.~Panwar, and N.~Goyal.
\newblock In-context learning through the bayesian prism.
\newblock \emph{arXiv preprint arXiv:2306.04891}, 2023.

\bibitem[Aky{\"u}rek et~al.(2022)Aky{\"u}rek, Schuurmans, Andreas, Ma, and Zhou]{akyurek2022learning}
E.~Aky{\"u}rek, D.~Schuurmans, J.~Andreas, T.~Ma, and D.~Zhou.
\newblock What learning algorithm is in-context learning? investigations with linear models.
\newblock \emph{arXiv preprint arXiv:2211.15661}, 2022.

\bibitem[Blei et~al.(2003)Blei, Ng, and Jordan]{blei2003latent}
D.~M. Blei, A.~Y. Ng, and M.~I. Jordan.
\newblock Latent dirichlet allocation.
\newblock \emph{Journal of machine Learning research}, 3\penalty0 (Jan):\penalty0 993--1022, 2003.

\bibitem[Brown et~al.(2020)Brown, Mann, Ryder, Subbiah, Kaplan, Dhariwal, Neelakantan, Shyam, Sastry, Askell, et~al.]{brown2020language}
T.~Brown, B.~Mann, N.~Ryder, M.~Subbiah, J.~D. Kaplan, P.~Dhariwal, A.~Neelakantan, P.~Shyam, G.~Sastry, A.~Askell, et~al.
\newblock Language models are few-shot learners.
\newblock \emph{Advances in neural information processing systems}, 33:\penalty0 1877--1901, 2020.

\bibitem[Chen et~al.(2024)Chen, Sheen, Wang, and Yang]{chen2024training}
S.~Chen, H.~Sheen, T.~Wang, and Z.~Yang.
\newblock Training dynamics of multi-head softmax attention for in-context learning: Emergence, convergence, and optimality.
\newblock \emph{arXiv preprint arXiv:2402.19442}, 2024.

\bibitem[Dai et~al.(2022)Dai, Sun, Dong, Hao, Ma, Sui, and Wei]{dai2022can}
D.~Dai, Y.~Sun, L.~Dong, Y.~Hao, S.~Ma, Z.~Sui, and F.~Wei.
\newblock Why can gpt learn in-context? language models implicitly perform gradient descent as meta-optimizers.
\newblock \emph{arXiv preprint arXiv:2212.10559}, 2022.

\bibitem[Devlin et~al.(2018)Devlin, Chang, Lee, and Toutanova]{devlin2018bert}
J.~Devlin, M.-W. Chang, K.~Lee, and K.~Toutanova.
\newblock Bert: Pre-training of deep bidirectional transformers for language understanding.
\newblock \emph{arXiv preprint arXiv:1810.04805}, 2018.

\bibitem[Edelman et~al.(2022)Edelman, Goel, Kakade, and Zhang]{edelman2022inductive}
B.~L. Edelman, S.~Goel, S.~Kakade, and C.~Zhang.
\newblock Inductive biases and variable creation in self-attention mechanisms.
\newblock In \emph{International Conference on Machine Learning}, pages 5793--5831. PMLR, 2022.

\bibitem[Fang et~al.(2024)Fang, Xu, Tan, Zhang, Hu, Qi, Nickleach, Socolinsky, Sengamedu, Faloutsos, et~al.]{fang2024large}
X.~Fang, W.~Xu, F.~A. Tan, J.~Zhang, Z.~Hu, Y.~J. Qi, S.~Nickleach, D.~Socolinsky, S.~Sengamedu, C.~Faloutsos, et~al.
\newblock Large language models (llms) on tabular data: Prediction, generation, and understanding-a survey.
\newblock 2024.

\bibitem[Fu et~al.(2023)Fu, Chen, Jia, and Sharan]{fu2023transformers}
D.~Fu, T.-Q. Chen, R.~Jia, and V.~Sharan.
\newblock Transformers learn higher-order optimization methods for in-context learning: A study with linear models.
\newblock \emph{arXiv preprint arXiv:2310.17086}, 2023.

\bibitem[Gruber et~al.(2007)Gruber, Weiss, and Rosen-Zvi]{gruber2007hidden}
A.~Gruber, Y.~Weiss, and M.~Rosen-Zvi.
\newblock Hidden topic markov models.
\newblock In \emph{Artificial intelligence and statistics}, pages 163--170. PMLR, 2007.

\bibitem[Han et~al.(2023)Han, Wang, Zhao, and Ji]{han2023explaining}
C.~Han, Z.~Wang, H.~Zhao, and H.~Ji.
\newblock Explaining emergent in-context learning as kernel regression.
\newblock 2023.

\bibitem[Hegselmann et~al.(2023)Hegselmann, Buendia, Lang, Agrawal, Jiang, and Sontag]{hegselmann2023tabllm}
S.~Hegselmann, A.~Buendia, H.~Lang, M.~Agrawal, X.~Jiang, and D.~Sontag.
\newblock Tabllm: Few-shot classification of tabular data with large language models.
\newblock In \emph{International Conference on Artificial Intelligence and Statistics}, pages 5549--5581. PMLR, 2023.

\bibitem[Huang et~al.(2023)Huang, Cheng, and Liang]{huang2023context}
Y.~Huang, Y.~Cheng, and Y.~Liang.
\newblock In-context convergence of transformers.
\newblock \emph{arXiv preprint arXiv:2310.05249}, 2023.

\bibitem[Jelodar et~al.(2019)Jelodar, Wang, Yuan, Feng, Jiang, Li, and Zhao]{jelodar2019latent}
H.~Jelodar, Y.~Wang, C.~Yuan, X.~Feng, X.~Jiang, Y.~Li, and L.~Zhao.
\newblock Latent dirichlet allocation (lda) and topic modeling: models, applications, a survey.
\newblock \emph{Multimedia tools and applications}, 78:\penalty0 15169--15211, 2019.

\bibitem[Kim and Suzuki(2024)]{kim2024transformers}
J.~Kim and T.~Suzuki.
\newblock Transformers learn nonlinear features in context: Nonconvex mean-field dynamics on the attention landscape.
\newblock \emph{arXiv preprint arXiv:2402.01258}, 2024.

\bibitem[Li et~al.(2023{\natexlab{a}})Li, Ildiz, Papailiopoulos, and Oymak]{li2023transformer}
Y.~Li, M.~E. Ildiz, D.~Papailiopoulos, and S.~Oymak.
\newblock Transformers as algorithms: Generalization and stability in in-context learning.
\newblock In \emph{International Conference on Machine Learning}, pages 19565--19594. PMLR, 2023{\natexlab{a}}.

\bibitem[Li et~al.(2023{\natexlab{b}})Li, Li, and Risteski]{li2023transformers}
Y.~Li, Y.~Li, and A.~Risteski.
\newblock How do transformers learn topic structure: Towards a mechanistic understanding.
\newblock In \emph{International Conference on Machine Learning}, pages 19689--19729. PMLR, 2023{\natexlab{b}}.

\bibitem[Li et~al.(2024)Li, Sreenivasan, Giannou, Papailiopoulos, and Oymak]{li2024dissecting}
Y.~Li, K.~Sreenivasan, A.~Giannou, D.~Papailiopoulos, and S.~Oymak.
\newblock Dissecting chain-of-thought: Compositionality through in-context filtering and learning.
\newblock \emph{Advances in Neural Information Processing Systems}, 36, 2024.

\bibitem[Mahankali et~al.(2023)Mahankali, Hashimoto, and Ma]{mahankali2023one}
A.~Mahankali, T.~B. Hashimoto, and T.~Ma.
\newblock One step of gradient descent is provably the optimal in-context learner with one layer of linear self-attention.
\newblock \emph{arXiv preprint arXiv:2307.03576}, 2023.

\bibitem[Min et~al.(2022)Min, Lewis, Zettlemoyer, and Hajishirzi]{min-etal-2022-metaicl}
S.~Min, M.~Lewis, L.~Zettlemoyer, and H.~Hajishirzi.
\newblock Metaicl: Learning to learn in context.
\newblock In M.~Carpuat, M.-C. de~Marneffe, and I.~V. Meza~Ruiz, editors, \emph{Proceedings of the 2022 Conference of the North American Chapter of the Association for Computational Linguistics: Human Language Technologies}, pages 2791--2809, Seattle, United States, Jul 2022. Association for Computational Linguistics.
\newblock \doi{10.18653/v1/2022.naacl-main.201}.
\newblock URL \url{https://aclanthology.org/2022.naacl-main.201}.

\bibitem[M{\"u}ller et~al.(2021)M{\"u}ller, Hollmann, Arango, Grabocka, and Hutter]{muller2021transformers}
S.~M{\"u}ller, N.~Hollmann, S.~P. Arango, J.~Grabocka, and F.~Hutter.
\newblock Transformers can do bayesian inference.
\newblock \emph{arXiv preprint arXiv:2112.10510}, 2021.

\bibitem[Olsson et~al.(2022)Olsson, Elhage, Nanda, Joseph, DasSarma, Henighan, Mann, Askell, Bai, Chen, et~al.]{olsson2022context}
C.~Olsson, N.~Elhage, N.~Nanda, N.~Joseph, N.~DasSarma, T.~Henighan, B.~Mann, A.~Askell, Y.~Bai, A.~Chen, et~al.
\newblock In-context learning and induction heads.
\newblock \emph{arXiv preprint arXiv:2209.11895}, 2022.

\bibitem[Radford et~al.(2019)Radford, Wu, Child, Luan, Amodei, Sutskever, et~al.]{radford2019language}
A.~Radford, J.~Wu, R.~Child, D.~Luan, D.~Amodei, I.~Sutskever, et~al.
\newblock Language models are unsupervised multitask learners.
\newblock \emph{OpenAI blog}, 1\penalty0 (8):\penalty0 9, 2019.

\bibitem[Ravent{\'o}s et~al.(2024)Ravent{\'o}s, Paul, Chen, and Ganguli]{raventos2024pretraining}
A.~Ravent{\'o}s, M.~Paul, F.~Chen, and S.~Ganguli.
\newblock Pretraining task diversity and the emergence of non-bayesian in-context learning for regression.
\newblock \emph{Advances in Neural Information Processing Systems}, 36, 2024.

\bibitem[Solaiman et~al.(2019)Solaiman, Brundage, Clark, Askell, Herbert-Voss, Wu, Radford, Krueger, Kim, Kreps, et~al.]{solaiman2019release}
I.~Solaiman, M.~Brundage, J.~Clark, A.~Askell, A.~Herbert-Voss, J.~Wu, A.~Radford, G.~Krueger, J.~W. Kim, S.~Kreps, et~al.
\newblock Release strategies and the social impacts of language models.
\newblock \emph{arXiv preprint arXiv:1908.09203}, 2019.

\bibitem[Van~der Maaten and Hinton(2008)]{van2008visualizing}
L.~Van~der Maaten and G.~Hinton.
\newblock Visualizing data using t-sne.
\newblock \emph{Journal of machine learning research}, 9\penalty0 (11), 2008.

\bibitem[Von~Oswald et~al.(2023)Von~Oswald, Niklasson, Randazzo, Sacramento, Mordvintsev, Zhmoginov, and Vladymyrov]{von2023transformers}
J.~Von~Oswald, E.~Niklasson, E.~Randazzo, J.~Sacramento, A.~Mordvintsev, A.~Zhmoginov, and M.~Vladymyrov.
\newblock Transformers learn in-context by gradient descent.
\newblock In \emph{International Conference on Machine Learning}, pages 35151--35174. PMLR, 2023.

\bibitem[Wang et~al.(2023)Wang, Zhu, Saxon, Steyvers, and Wang]{wang2023ICL}
X.~Wang, W.~Zhu, M.~Saxon, M.~Steyvers, and Y.~W. Wang.
\newblock Large language models are latent variable models: Explaining and finding good demonstrations for in-context learning.
\newblock \emph{arXiv preprint arXiv:2301.11916}, 2023.

\bibitem[Wei et~al.(2022)Wei, Wang, Schuurmans, Bosma, Xia, Chi, Le, Zhou, et~al.]{wei2022chain}
J.~Wei, X.~Wang, D.~Schuurmans, M.~Bosma, F.~Xia, E.~Chi, Q.~V. Le, D.~Zhou, et~al.
\newblock Chain-of-thought prompting elicits reasoning in large language models.
\newblock \emph{Advances in neural information processing systems}, 35:\penalty0 24824--24837, 2022.

\bibitem[Wu et~al.(2023)Wu, Zou, Chen, Braverman, Gu, and Bartlett]{wu2023many}
J.~Wu, D.~Zou, Z.~Chen, V.~Braverman, Q.~Gu, and P.~L. Bartlett.
\newblock How many pretraining tasks are needed for in-context learning of linear regression?
\newblock \emph{arXiv preprint arXiv:2310.08391}, 2023.

\bibitem[Xie et~al.(2021)Xie, Raghunathan, Liang, and Ma]{xie2021explanation}
S.~M. Xie, A.~Raghunathan, P.~Liang, and T.~Ma.
\newblock An explanation of in-context learning as implicit bayesian inference.
\newblock \emph{arXiv preprint arXiv:2111.02080}, 2021.

\bibitem[Xing et~al.(2024)Xing, Lin, Suh, Song, and Cheng]{xing2024benefits}
Y.~Xing, X.~Lin, N.~Suh, Q.~Song, and G.~Cheng.
\newblock Benefits of transformer: In-context learning in linear regression tasks with unstructured data.
\newblock \emph{arXiv preprint arXiv:2402.00743}, 2024.

\bibitem[Yang et~al.(2022)Yang, Wipf, et~al.]{yang2022transformers}
Y.~Yang, D.~P. Wipf, et~al.
\newblock Transformers from an optimization perspective.
\newblock \emph{Advances in Neural Information Processing Systems}, 35:\penalty0 36958--36971, 2022.

\bibitem[Zhang et~al.(2023{\natexlab{a}})Zhang, Frei, and Bartlett]{zhang2023trained}
R.~Zhang, S.~Frei, and P.~L. Bartlett.
\newblock Trained transformers learn linear models in-context.
\newblock \emph{arXiv preprint arXiv:2306.09927}, 2023{\natexlab{a}}.

\bibitem[Zhang et~al.(2023{\natexlab{b}})Zhang, Zhang, Yang, and Wang]{zhang2023and}
Y.~Zhang, F.~Zhang, Z.~Yang, and Z.~Wang.
\newblock What and how does in-context learning learn? bayesian model averaging, parameterization, and generalization.
\newblock \emph{arXiv preprint arXiv:2305.19420}, 2023{\natexlab{b}}.

\end{thebibliography}
\bibliographystyle{abbrvnat}
\newpage

\newpage
\onecolumn
\appendix

%
\hyphenation{op-tical net-works semi-conduc-tor}

%

%
%


%
\section{Justification of modeling and assumptions}\label{sec:gene}
\begin{rmk}\label{rmk:data}
The above modeling of sequence generation effectively captures key characteristics of natural language generation, ensuring both coherence and factual alignment. 1) Topic Consistency: Usually $p_t(\cdot|\theta)$ assigns higher probability to topics related to $\theta$; 2) The class attributes of key tokens align with real-world facts, as determined by the first token. In addition, it is important to highlight that unrelated topics—such as `Occupation' and `Geography'—are unlikely to occur under a shared concept. Even if two words belong to the same class (e.g, `Trump' and `Earthquake'), they are not likely to appear together in the same sequence.
\end{rmk}
\begin{rmk}\label{rmk:promptcompare}
(Justifying Assumption \ref{ass:example})
In Assumption \ref{ass:example}, we make two relatively strong assumptions, i.e (i) and (ii). We clarify that (i) and (ii) are reasonable in our illustrated example to show the difference in the predictions with \eqref{eq:prompt} and \eqref{eq:fakeprompt}. (i) is commonly used in literature \citep{huang2023context,zhang2023trained}, where the masked output is denoted as $\mathbf{0}$ (which is showed in \eqref{eq:fakeprompt}). (ii) assumes uniform attention. However, notice that for general $\boldsymbol{W}^Q$ and $\boldsymbol{W}^K$, it only changes the coefficient of the embedded context samples and $X_{\text q}$ in \eqref{eq:truepred}, and the coefficient of $y_i$ in \eqref{eq:fakepred}. More specifically, for the term $\boldsymbol{W}^V(X_i+y_i)$ in \eqref{eq:truepred}, the coefficient is some constant rather than $\frac{1}{2n+1}$; in \eqref{eq:fakepred}, the coefficients of $y_i$ is some constant instead of $\frac{1}{n+1}$. This does not change the fundamental difference between \eqref{eq:truepred} and \eqref{eq:fakepred}. %
\end{rmk}

\begin{rmk}\label{rmk:sequence}
Assumption \ref{ass:attention} characterizes the correlation (quantified by attention $A(\cdot)$) between the masked words $({(\widetilde{X}_{\text q})}_{:,j},\;L_1+1\leqslant j\leqslant N)$ and the previous words with the attention head. Intuitively, when a word is far from the masked word in a sequence, it contributes less to the masked word in prediction. We model this position factor by an ascending sequence of weights $a_1<a_2<\cdots<a_{n+1}$. For a sequence with a length larger than $N$, we assume the correlation between the word and the masked word will decay with the ``distance'' increasing (thus $a_i<a_i',\forall i<i'$). To simplify our analysis, we assume the token correlation within length $N$ be a uniform matrix (with each entry in $A(\cdot)$ equal to $\frac{1}{N}$). In Appendix \ref{sec:synthetic}, we empirically verify that if we freeze uniform attention and only update $\boldsymbol{W}^V$ in the training phase, the performance is very close to updating all the variables. A similar observation is found in \citep{li2023transformers}, where most theoretical results are also based on uniform attention.
\end{rmk}
\begin{figure}[h]
    \centering
    \includegraphics[width=0.8\linewidth]{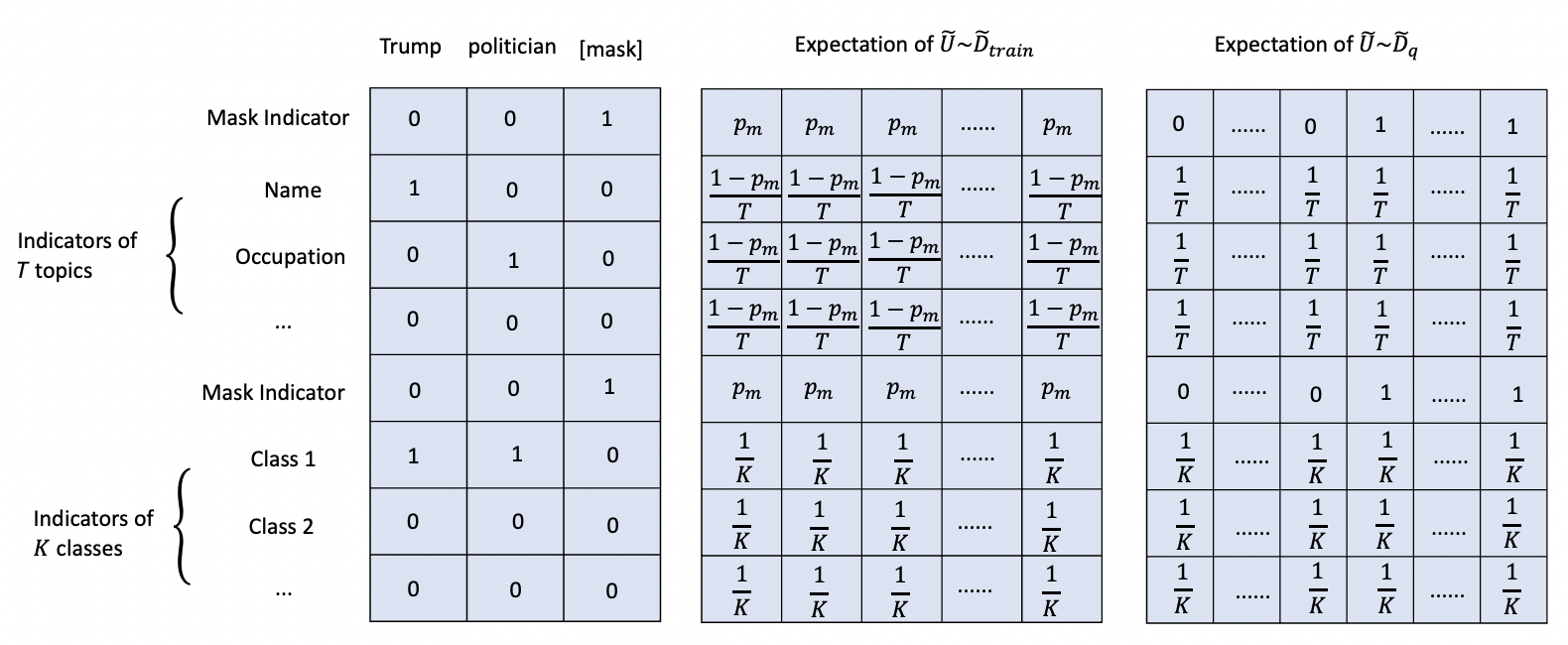}
    \caption{Example of encoded sequence.}
    \label{fig:tableexample}
\end{figure}
\section{Proof of Theorem \ref{thm:bayes}}
\begin{proof}
We can derive the prediction probability of label $y$ as: 
\begin{align*}
&\quad p\left(y \mid R_1, R_2,\cdots, R_H,S_n, X_{\text q}\right)\\
&=\int_\theta p\left(y \mid R_1, R_2,\cdots, R_H,S_n, X_{\text q}, \theta\right) p\left(\theta \mid R_1,R_2,\cdots,R_H, S_n, X_{\text q}\right) d \theta\\
&\propto \int_{\theta} p\left(y \mid R_1,R_2,\cdots, R_H,S_n, X_{\text q},\theta\right) p\left(R_1, R_2,\cdots, R_H,S_n, X_{\text q}\mid \theta\right) p(\theta) d \theta\\
&\propto \int_{\theta}p\left(y \mid S_n, X_{\text q},\theta\right)\cdot \frac{p\left(R_1,R_2,\cdots,R_H,S_n, X_{\text {q}} \mid \theta\right)}{p\left(R_1,R_2,\cdots,R_H,S_n, X_{\text {q}} \mid \theta^*\right)}\cdot p\left(\theta\right)d \theta\\
&\propto  \int_{\theta}p\left(y \mid S_n, X_{\text q},\theta\right)\cdot \frac{p\left(R_1,R_2,\cdots,R_H\mid \theta\right)}{p\left(R_1,R_2,\cdots,R_h\mid \theta^*\right)}\cdot \frac{p\left(S_n,X_{\text q}\mid \theta\right)}{p\left(S_n,X_{\text q}\mid \theta^{*}\right)}\cdot p\left(\theta\right)d \theta\\
\end{align*}
Next, we aim to show that
\begin{align*}
\frac{p\left(R_1,R_2,\cdots,R_H\mid \theta\right)}{p\left(R_1,R_2,\cdots,R_H\mid \theta^*\right)}\cdot \frac{p\left(S_n,X_{\text q}\mid \theta\right)}{p\left(S_n,X_{\text q}\mid \theta^{*}\right)}\stackrel{n\rightarrow\infty}\longrightarrow 0,\; \forall \theta\neq \theta^{*}.
\end{align*}
Define $r_{n_1}(\theta):=\frac{1}{n_1H}\sum\limits_{h=1}^{H}\sum\limits_{i=1}^{n_1}\log\frac{p\left(R_{h,i}\mid \theta\right)}{p\left(R_{h,i}\mid \theta^{*}\right)}$, $q_n(\theta):=\frac{1}{n}\sum\limits_{i=1}^{n}\log\frac{p\left(s_i\mid \theta\right)}{p\left(s_i\mid \theta^{*}\right)}$.
Then we have
\begin{align*}
\frac{p\left(R_1,R_2,\cdots,R_H\mid \theta\right)}{p\left(R_1,R_2,\cdots,R_H\mid \theta^*\right)}\cdot \frac{p\left(S_n,X_{\text q}\mid \theta\right)}{p\left(S_n,X_{\text q}\mid \theta^{*}\right)}=\exp\left(n_1 H\cdot r_{n_1}(\theta)+n\cdot q_n(\theta)\right)  
\end{align*}
(1) With probability$>0.999$, $r_{n_1}(\theta)<0$.
We know $\forall i \in[n_1]$,
\begin{align*}
&\mathbb{E}[r_{n_1}(\theta)]=\mathbb{E}_{R_{h,i}\sim p(\cdot\mid \theta_h)}\left[\frac{1}{n_1 T} \sum_{h=1}^H \sum_{i=1}^{n_1} \log \frac{p\left(R_{h, i} \mid \theta\right)}{p\left(R_{h, i} \mid \theta^*\right)}\right]\\
&=\frac{1}{H}\sum_{h=1}^H\mathbb{E}_{R_{h,i}\sim p(\cdot\mid \theta_h)}\left[ \log \frac{p\left(R_{h, i} \mid \theta\right)}{p\left(R_{h, i} \mid \theta^*\right)}\right]\\
&=\frac{1}{T}\sum_{h=1}^H\mathbb{E}_{R_{h,i}\sim p(\cdot\mid \theta_h)}\left[\log \frac{p\left(R_{h, i} \mid \theta\right)}{p\left(R_{h, i} \mid \theta_h\right)}+\log \frac{p\left(R_{h, i} \mid \theta_h\right)}{p\left(R_{h, i} \mid \theta^*\right)}\right]\\
&=\frac{1}{T}\sum\limits_{t=1}^{T}\text{KL}\left(p\left(\cdot\mid\theta_h\right)||p\left(\cdot\mid\theta^{*}\right)\right)-\text{KL}\left(p\left(\cdot\mid\theta_h\right)||p\left(\cdot\mid\theta\right)\right)
\end{align*}
By Assumption, we have
\begin{align*}
\frac{1}{T} \sum_{h=1}^H \mathrm{KL}\left(p\left(\cdot \mid \theta_h^*\right) \| p\left(\cdot \mid \theta^*\right)\right)-\operatorname{KL}\left(p\left(\cdot \mid \theta_h\right) \| p(\cdot \mid \theta)\right):=c_1<0
\end{align*}
In addition, By Assumption \ref{ass:bound_variance}, the variance of $r_{n_1}(\theta)$ can be bounded as following:
\begin{align*}
&\operatorname{var}\left(r_{n_1}(\theta)\right) = \operatorname{var}\left(\frac{1}{n_1 T} \sum_{h=1}^H \sum_{i=1}^{n_1} \log \frac{p\left(R_{h, i} \mid \theta\right)}{p\left(R_{h, i} \mid \theta^*\right)}\right)\leqslant\frac{\sigma^2}{n_1 H}
\end{align*}
Then by Central Limit Theorem (CLT), with probability $>0.999$,
\begin{align*}
r_{n_1}(\theta)<c_1+\frac{3\sigma}{\sqrt{n_1H}}:=c_1'<0
\end{align*}

(2) With probability $>0.999$, $q_n(\theta)>0$. Similar to the derivation in (1), we have
\begin{align*}
q_n(\theta)\leqslant -KL\left(p\left(\cdot\mid \theta^{*}\right)||p\left(\cdot\mid \theta\right)\right)+\frac{3\sigma}{\sqrt{n}}:=c_2'<0.
\end{align*}
\begin{align*}
\frac{p\left(R_1,R_2,\cdots,R_H\mid \theta\right)}{p\left(R_1,R_2,\cdots,R_H\mid \theta^*\right)}\cdot \frac{p\left(S_n,X_{\text q}\mid \theta\right)}{p\left(S_n,X_{\text q}\mid \theta^{*}\right)}=\exp\left(n_1 H\cdot r_{n_1}(\theta)+n\cdot q_n(\theta)\right)<\epsilon 
\end{align*}
Then we can derive
\begin{align*}
-(n_1Hc_1'+nc_2')>\log\frac{1}{\epsilon}.
\end{align*}
Suppose holds, then we have the following equation:
\begin{align*}
&\int_\theta p\left(y \mid S_n, X_{\text {q}}, \theta\right) \cdot \frac{p\left(R_1, R_2, \cdots, R_H \mid \theta\right)}{p\left(R_1, R_2, \cdots, R_H \mid \theta^*\right)} \cdot \frac{p\left(S_n, X_{\text {q}} \mid \theta\right)}{p\left(S_n, X_{\text {q}} \mid \theta^*\right)} \cdot p(\theta) d \theta\\
&=p\left(y \mid S_n, X_{\text {q}}, \theta^{*}\right)p\left(\theta^{*}\right)+\int_{\theta\neq\theta^{*}}p\left(y \mid S_n, X_{\text {q}}, \theta\right)\cdot\exp \left(n_1 H \cdot r_{n_1}(\theta)+n \cdot q_n(\theta)\right)\cdot p(\theta) d \theta\\
&\propto p\left(y \mid S_n, X_{\text {q}}, \theta^{*}\right)+\frac{1}{p(\theta^{*})}\int_{\theta\neq\theta^{*}}p\left(y \mid S_n, X_{\text {q}}, \theta\right)\cdot\exp \left(n_1 H \cdot r_{n_1}(\theta)+n \cdot q_n(\theta)\right)\cdot p(\theta) d \theta\\
&= p\left(y \mid S_n, X_{\text {q}}, \theta^{*}\right)+\frac{1}{p(\theta^{*})}\int_{\theta\neq\theta^{*}}\epsilon_{\theta}(y)\cdot p(\theta) d\theta
\end{align*}
Thus, we can conclude that 
\begin{align*}
p\left(y \mid R_1, R_2, \cdots, R_h, S_n, X_{\text {q}}\right)\propto p\left(y \mid S_n, X_{\text {q}}, \theta^*\right)+\frac{1}{p\left(\theta^*\right)} \int_{\theta \neq \theta^*} \epsilon_\theta(y) \cdot p(\theta) d \theta,\;\forall y.
\end{align*}
Since we know
\begin{align*}
c\sum\limits_{y\in \mathcal{Y}}p\left(y \mid S_n, X_{\text {q}}, \theta^*\right)+\frac{1}{p\left(\theta^*\right)} \int_{\theta \neq \theta^*} \epsilon_\theta(y) \cdot p(\theta) d \theta=1
\end{align*}
We have $c = 1/\sum_{y \in \mathcal{Y}} p\left(y \mid S_n, X_{\text {q}}, \theta^*\right)+\frac{1}{p\left(\theta^*\right)} \int_{\theta \neq \boldsymbol{\theta}^*} \epsilon_\theta(y) \cdot p(\theta) d \theta$. Let $y^*=\operatorname{argmax}_{y\in\mathcal{Y}}p\left(y\mid S_n,X_{\text q},\theta\right)$
\begin{align*}
&p\left(y^{*} \mid R_1, R_2, \cdots, R_H, S_n, X_{\text {q}}\right)=c\cdot p\left(y^* \mid S_n, X_{\text {q}}, \theta^*\right)+\frac{1}{p\left(\theta^*\right)} \int_{\theta \neq \boldsymbol{\theta}^*} \epsilon_\theta(y) \cdot p(\theta) d \theta\\
&\geqslant c\cdot p\left(y^* \mid S_n, X_{\text {q}}, \theta^*\right)\\
&>c\cdot\left(\max\limits_{y\neq y^*}p\left(y \mid S_n, X_{\text {q}}, \theta^*\right)+\frac{\epsilon}{p(\theta^*)}\right)
\end{align*}
Thus, we can conclude that $\operatorname{argmax}_{y\in\mathcal{Y}}p\left(y \mid R_1, R_2, \cdots, R_H, S_n, X_{\text {q}}\right)=y^*$

\end{proof}

\section{Proof of Claim \ref{thm:lda}}\label{sec:proofclaim}
(1) \textbf{Prediction without ICL:} This is computed directly by$$\widehat{y}_{\text q}=f_{\bf w}(\widetilde{U}_{\text {q}})_{1:D,L_1+1:N}$$

(2) \textbf{Prediction with ICL:} Given additional $n$ encoded context samples $S_n=s_{1:n}$, with encoded matrix $U_i\in{\{0,1\}^{(T+K+2) \times N}}\sim \mathcal{D}_{\text q}$. We construct the encoded prompt by \eqref{eq:prompt}, which can be written as: $U_{\text{stacked}}=(U_{1:n}, \widetilde{U}_{\text q})$,
where the last $L_2$ columns in $\widetilde{U}_{\text q}$ represent masks. We predict the masks as $\widehat{y}_{\text q}=f_{\bf w}\left(\mathbf{Z}\right)_{1:D,nN+L_1+1:N}$.

\begin{claim}
(Formal Version) Generate each encoded pre-train data $U_{\text{train}}\sim\mathcal{D}_{\text{train}},\widetilde{U}_{\text{train}}\sim\widetilde{\mathcal{D}}_{\text{train}}$. There exists $a_1<a_2<\cdots<a_{n+1}$, such that if we train a one-layer Transformer \eqref{eq:transformer}: (1) With attention $A(\cdot)$ that satisfies Assumption \ref{ass:attention}; (2) By minimizing the $\ell_2$-regularized objective function \eqref{eq:obj} with variable $\boldsymbol{W}^{V}$. Suppose $\boldsymbol{W}^{{V}*}\in \underset{\lambda\rightarrow 0}{\operatorname{lim}}\operatorname{argmax}L(\boldsymbol{W}^V)$, and we use $f_{W^{\boldsymbol{V}*}}(\cdot)$ as prediction model. Given a masked query sequence $\widetilde{X}_\text{q}$ encoded as $\widetilde{U}_\text{q}$, predict the last $L_2$ tokens ($L_1+1\leqslant j\leqslant N$) as $\widehat{y}_q$. We have following :\\
\begin{align*}
(1)\begin{cases}
&f_{\bf w}(\widetilde{U}_{\text {q}})_{l,j}=1/T,\;l=1,2,\cdots,T,\\
&\underset{T+1\leqslant l\leqslant T+K+1}{\arg\max}f_{\bf w}(\widetilde{U}_{\text {q}})_{l,j}=T+k^{*}_{\text q}+1
\end{cases}
\end{align*}
\begin{align*}
(2)\begin{cases}
&\underset{0\leqslant l\leqslant T}{\arg\max}\ f_{\bf w}({U}_{\text{stacked}})_{l,j}=t^{*}\\
&\underset{T+1\leqslant l\leqslant T+K+1}{\arg\max}f_{\bf w}({U}_{\text{stacked}})_{l,j}=T+k^{*}_{\text q}+1
\end{cases}
\end{align*}
\end{claim}

\begin{proof}
First, let us derive the closed form of ${\boldsymbol{W}}^{V*}$. The proof idea follows from \citep{li2023transformers}. For a document $\boldsymbol{w}$, when the document length $N$ is large, for $i=0,1,\cdots,T+K+1,j\geqslant2$. Define $\mathbf{1}(\cdot)$ as the indicator function. We have the following equation:
\begin{align*}
&[\widetilde{{X}} A(\widetilde{U})]_{l j}=\frac{1}{N} \sum_{p=1}^N \widetilde{U}_{lp}=\frac{1}{N} \sum_{p=1}^N \mathbf{1}_{\widetilde{U}_{l p}=1}\\
&=\begin{cases}p_m & \text { if } l=0 \\ P_{\boldsymbol{w}}(l)\left(1-p_m\right) & \text { if } l \in[T]\\
p_m&\text{ if } l=T+1\\
Q&\text{ if } l=T+k^{*}+1\\
\frac{1-Q}{K-1}&\text{ if }T+2\leqslant l\leqslant T+K+1,l\neq T+k^{*}+1\end{cases}
\end{align*}
The prediction of the $j$-th column can be written as following:
\begin{align*}
&\quad(\boldsymbol{W} \widetilde{{U}} A(\widetilde{{U}}))_{l j}\\
&=\begin{cases}\boldsymbol{W}_{l0}p_m+\sum\limits_{r=1}^{T}\boldsymbol{W}_{lr}P_{\boldsymbol{w}}(r)(1-p_m)&\text{ if }l\in [T]\\
\boldsymbol{W}_{l0}p_m+\sum\limits_{l=1}^{T}\boldsymbol{W}_{lr}P_{\boldsymbol{w}}(r\mid X_1)(1-p_m)&\text{ if }T+1\leqslant l\leqslant T+K+1
\end{cases}
\end{align*}
Recall the loss function is
\begin{align}
L(\boldsymbol{W})=\mathbb{E}_{U \sim \mathcal{D}_{\text{train}}} \frac{1}{N} \sum_{j=1}^N\left\|(\boldsymbol{W} \widetilde{U} A(\widetilde{U}))_{: j}-X_{: j}\right\|_2^2
\end{align}
It is easy to show that $\forall U$, $L(\boldsymbol{W})$ is minimized at 
\begin{align*}
(\boldsymbol{W} \widetilde{U} A(\widetilde{U}))_{: j}=\frac{1}{N} \sum_{p=1}^N U_{: p},
\end{align*}
which is equivalent to
\begin{align*}
\begin{cases}
&(\boldsymbol{W} \widetilde{U} A(\widetilde{U}))_{0 j} =0 \\
&(\boldsymbol{W} \widetilde{U} A(\widetilde{U}))_{l j} =P_{\boldsymbol{w}}(i), \quad i \in [T]\\
&(\boldsymbol{W} \widetilde{U} A(\widetilde{U}))_{T+1, j} =0. \\
&(\boldsymbol{W} \widetilde{U} A(\widetilde{U}))_{l j} =P_{\boldsymbol{w}}(l\mid U_1), \quad  T+2\leqslant l\leqslant T+K+1
\end{cases}
\end{align*}
(1) Consider the first $T+1$ rows.
\begin{align}\label{eq:topicequation}
\begin{cases}\boldsymbol{W}_{00}p_m+\sum\limits_{r=1}^{T}\boldsymbol{W}_{0l}P_{\boldsymbol{w}}(r)(1-p_m)=0\\
\boldsymbol{W}_{l0}p_m+\sum\limits_{r=1}^{T}\boldsymbol{W}_{lr}P_{\boldsymbol{w}}(r)(1-p_m)=P_{\boldsymbol{w}}(l)\text{ if }l\in [T].
\end{cases}  
\end{align}
We claim that $\forall l$, there exists $ u_l$ such that $\forall r\neq l$, $\boldsymbol{W}_{lj}=u_l$.
We prove the claim by contradiction. Suppose the above claim does not hold. We consider a special case where topic $`1'$ is selected topic in Step 1 in data generation Section \ref{sec:data}. Then the following equations hold:
\begin{align*}
\begin{cases}
&P_{\boldsymbol{w}}(1)=\frac{1}{2}=\boldsymbol{W}_{1 0} p_m+\boldsymbol{W}_{1 2}\cdot\frac{1}{2}\cdot(1-p_m),\quad\text{if $`1'$ and $`2'$ are selected}, \\
&P_{\boldsymbol{w}}(1)=\frac{1}{2}=\boldsymbol{W}_{1 0} p_m+\boldsymbol{W}_{1 3}\cdot\frac{1}{2}\cdot(1-p_m),\quad\text{if $`1'$ and $`3'$ are selected}.
\end{cases}
\end{align*}
Suppose the above two equations need to hold, then there is $\boldsymbol{W}_{12}=\boldsymbol{W}_{13}=u_1$. Similarly, $\forall l$, there exists $u_l=\boldsymbol{W}_{lr},\; r\neq l$. Thus, \eqref{eq:topicequation} can be written as:
\begin{align*}
\begin{cases}
&\boldsymbol{W}_{00}=-\frac{u_0(1-p_m)}{p_m}\\
&\boldsymbol{W}_{l0}p_m+u_l(1-p_m)+P_{\boldsymbol{w}}(l)\left(\boldsymbol{W}_{ll}(1-p_m)-u_l(1-p_m)-1\right)=0
\end{cases}
\end{align*}
Then we have
\begin{align*}
\begin{cases}
&\boldsymbol{W}_{00}=-\frac{u_0(1-p_m)}{p_m}\\
&\boldsymbol{W}_{ll}=u_l+\frac{1}{1-p_m}\\
&\boldsymbol{W}_{l0}=-\frac{u_l(1-p_m)}{p_m}
\end{cases}    
\end{align*}
(2) Consider the last $K+1$ rows.
\begin{align}\label{eq:classequation}
\begin{cases}
&\boldsymbol{W}_{T+1,0}p_m+\sum\limits_{r=T+1}^{T+K+1}\boldsymbol{W}_{T+1,l}P_{\boldsymbol{w}}(r\mid U_1)\left(1-p_m\right)=0\\
&\boldsymbol{W}_{l0}p_m+\sum\limits_{r=T+2}^{T+K+1}\boldsymbol{W}_{lr}P_{\boldsymbol{w}}(l\mid U_1)(1-p_m)=P_{\boldsymbol{w}}(l\mid U_1)
\end{cases}
\end{align}
Now we claim, $\forall r\neq l$, there exists $q_l$ such that $\boldsymbol{W}_{lr}=q_l$. We prove the claim by contradiction. Suppose $l\neq k^{*}$, then $\forall Q$, we have the following equation:
\begin{align*}
\begin{cases}
&\frac{1-Q}{K-1}=\boldsymbol{W}_{l 0} p_m+\boldsymbol{W}_{ll_1}Q(1-p_m)+\sum\limits_{r\neq l_1}\boldsymbol{W}_{rl_1}\cdot \frac{1-Q}{K-1}(1-p_m),\quad \text{$\text{key class}=l_1$}\\
&\frac{1-Q}{K-1}=\boldsymbol{W}_{l 0} p_m+\boldsymbol{W}_{ll_2}Q(1-p_m)+\sum\limits_{r\neq l_2}\boldsymbol{W}_{rl_2}\cdot \frac{1-Q}{K-1}(1-p_m),\quad \text{$\text{key class}=l_2$}.
\end{cases}
\end{align*}
Thus, we must have $\boldsymbol{W}_{ll_1}=\boldsymbol{W}_{ll_2}=q_l$. We can rewrite \eqref{eq:classequation} as following:
\begin{align*}
\begin{cases}
&\boldsymbol{W}_{T+1,0} p_m+q_{T+1}(1-p_m)=0\\
&\boldsymbol{W}_{l 0} p_m+\boldsymbol{W}_{ll}P_{\boldsymbol{w}}(l \mid X_1)(1-p_m)+q_l\left(1-P_{\boldsymbol{w}}(l \mid U_1)\right)(1-p_m)=P_{\boldsymbol{w}}\left(l \mid U_1\right)
\end{cases}
\end{align*}
Then we have
\begin{align*}
\begin{cases}
&\boldsymbol{W}_{T+1,0}=-\frac{q_{T+2}\left(1-p_m\right)}{p_m}\\
&\boldsymbol{W}_{ll}=q_i+\frac{1}{1-p_m}\\
&\boldsymbol{W}_{l 0}=-\frac{q_l\left(1-p_m\right)}{p_m}
\end{cases}
\end{align*}
By Lemma \ref{lemma:S}, it suffices to find $u_l,q_l$ that minimizes $\|\boldsymbol{W}^V\|_F$.
Now consider the first $T+1$ rows of $\boldsymbol{W}^{V}$.
\begin{align*}
&\|\boldsymbol{W}^V_{0:T,:}\|_F^2=\frac{u_0^2\left(1-p_m\right)^2}{p_m^2}+Tu_0^2+\sum\limits_{l=1}^{T}\frac{u_l^2\left(1-p_m\right)^2}{p_m^2}+(T-1)u_l^2+\left(\frac{1}{1-p_m}+u_l\right)^2
\end{align*}
It is easy to derive that $u_0^{*}=0$. For $l=1,2,\cdots,T$, we take the derivative
\begin{align*}
&\quad\frac{\partial \|\boldsymbol{W}_{l,:}\|^2_F}{\partial u_l}=2\left(\frac{u_l(1-p_m)^2}{p_m^2}+(T-1)u_l+u_l+\frac{1}{1-p_m}\right)\\
&=2\left(\left(T+\frac{(1-p_m)^2}{p_m^2}\right)u_l+\frac{1}{1-p_m}\right)=0
\end{align*}
So we derive 
\begin{align}\label{eq:optimalt}
\begin{cases}
&u_l=-\frac{1}{(1-p_m)\cdot\left(T+\frac{(1-p_m)^2}{p_m^2}\right)}=u^{*}\\
&\boldsymbol{W}_{ll}=u_i+\frac{1}{1-p_m}=u^{*}+\frac{1}{1-p_m}\\
&\boldsymbol{W}_{l0}=\frac{-u_l(1-p_m)}{p_m}=\frac{-u^{*}(1-p_m)}{p_m}
\end{cases}
\end{align}
Similarly, for the last $K+1$ rows, we have
\begin{align}\label{eq:optimalk}
\begin{cases}
&q_l=\frac{1}{\left(1-p_m\right) \cdot\left(K+\frac{\left(1-p_m\right)^2}{p_m^2}\right)}=q^*\\
&\boldsymbol{W}_{ll}=q_l+\frac{1}{1-p_m}=q^*+\frac{1}{1-p_m}\\
&\boldsymbol{W}_{l 0}=\frac{-q_l\left(1-p_m\right)}{p_m}=\frac{-q^*\left(1-p_m\right)}{p_m}
\end{cases}
\end{align}
Up to now we have characterized the optimal solution $\boldsymbol{W}^{V*}$. In the following part, we will compare the prediction with or without the prompt.
Given input $\widetilde{X}_{\text q}$, recall that our goal is to predict the last $L_2$ masked columns in $\widetilde{U}_{\text q}$. For $\forall L_1+1\leqslant j\leqslant N$ we have the following equation:
\begin{align}
&f_{\bf w}\left(\widetilde{U}_{\text {q}}\right)_{:,j}=\boldsymbol{W}^{V*}\cdot\widetilde{U}_{\text q}\cdot \left(A(\widetilde{U}_{\text q})\right)_{:,j}\nonumber\\
&=\boldsymbol{W}^{V*}\cdot \widetilde{U}_{\text q}\cdot\left(\frac{1}{N},\cdots,\frac{1}{N}\right)^{\top}\nonumber\\
&=\boldsymbol{W}^{V*}\cdot\left(\frac{1}{N}\sum\limits_{p=1}^{N}\boldsymbol{1}_{\left({\widetilde{U}}_{\text q}\right)_{0p}=1},\cdots,\frac{1}{N}\sum\limits_{p=1}^{N}\boldsymbol{1}_{\left({\widetilde{U}}_{\text q}\right)_{T+K+1,p}=1}\right)^{\top}\label{eq:querypred}
\end{align}
First, let us consider the first $T+1$ rows in the above prediction \eqref{eq:querypred}, which is the the topic predictor. Since we consider the case where the sequence length $N$ goes infinity, we have
\begin{align*}
f_{\bf w}\left(\widetilde{U}_{\text {q}}\right)_{0:T, j}&=\boldsymbol{W}^{V*}\cdot\left(\frac{1}{N}\sum\limits_{p=1}^{N}\boldsymbol{1}_{\left({\widetilde{U}}_{\text q}\right)_{0p}=1},\cdots,\frac{1}{N}\sum\limits_{p=1}^{N}\boldsymbol{1}_{\left({\widetilde{U}}_{\text q}\right)_{T+K+1,p}=1}\right)_{0:T}^{\top}\\
&=\boldsymbol{W}^{V*}\cdot\left(p_m,\frac{1-p_m}{T},\cdots,\frac{1-p_m}{T}\right)^{\top}\in\mathbb{R}^{T+1}
\end{align*}
Recall the optimal $\boldsymbol{W}^{V*}$. For $ l\in[T]$,
\begin{align*}
&\boldsymbol{W}^{V*}\cdot\left(p_m,\frac{1-p_m}{T},\cdots,\frac{1-p_m}{T}\right)_{l}^{\top}\\
&=\frac{-u^{*}(1-p_m)}{p_m}\cdot p_m+\left(u^{*}+\frac{1}{1-p_m}\right)\cdot\frac{1-p_m}{T}+(T-1)u^{*}\cdot\frac{1-p_m}{T}\\
&=\frac{1}{T} 
\end{align*}
The result implies that the predicted topics have the same probability. 

Second, we consider the class predictor in \eqref{eq:querypred}, which is the last $K+1$ rows in \eqref{eq:querypred}. For simplicity, we consider the case where the key word class $k^{*}_{\text q}=1$.
\begin{align*}
f_{\bf w}\left(\widetilde{U}_{\text {q}}\right)_{T+1:T+K+1,j}&=\boldsymbol{W}^{V*}\cdot\left(\frac{1}{N}\sum\limits_{p=1}^{N}\boldsymbol{1}_{\left({\widetilde{U}}_{\text q}\right)_{0p}=1},\cdots,\frac{1}{N}\sum\limits_{p=1}^{N}\boldsymbol{1}_{\left({\widetilde{U}}_{\text q}\right)_{T+K+1,p}=1}\right)_{T+1:T+K+1}^{\top}\\
&=\boldsymbol{W}^{V*}\cdot\left(p_m,Q(1-p_m),\frac{1-Q}{K-1}(1-p_m),\cdots\frac{1-Q}{K-1}(1-p_m)\right)^{\top}
\end{align*}
For $l=T+k^{*}_{\text q}+1=T+K+2$, we have
\begin{align*}
f_{\bf w}\left(\widetilde{U}_{\text {q }}\right)_{l, j}&=\frac{-v^*\left(1-p_m\right)}{p_m}\cdot p_m+\left(v^{*}+\frac{1}{1-p_m}\right)\cdot Q(1-p_m)+(K-1)\cdot v_i\cdot\frac{1-Q}{K-1}\left(1-p_m\right)\\
&=-v^*\left(1-p_m\right)+v^{*}\cdot Q(1-p_m)+Q+v^{*}(1-Q)(1-p_m)\\
&=Q
\end{align*}
Similarly, we can compute that for $l\neq T+K+2$, we have 
\begin{align*}
f_{\bf w}\left(\widetilde{U}_{\text {q}}\right)_{l, j}=\frac{1-Q}{K-1}. 
\end{align*}

Next, we consider the case where we have the in-context samples $s_1,s_2,\cdots,s_n$.
\begin{align}
f_{\bf w}\left(U_{\text{stacked}}\right)_{:,j}&=\boldsymbol{W}^{V*}\cdot U_{\text{stacked}}\cdot \left(A(U_{\text{stacked}})\right)_{:,j}\nonumber\\
&=\boldsymbol{W}^{V*}\cdot\left(U_1,\cdots,U_n,\widetilde{U}_{\text q}\right)\cdot\left(a_1/N,\cdots,a_1/N,a_2/N,\cdots,a_2/N,\cdots,a_{n+1}/N,\cdots,a_{n+1}/N\right)^{\top}\label{eq:promptpred}
\end{align}
Similarly, we consider the first $T+1$ entries in \eqref{eq:promptpred}, which is the topic predictor. Recall the query distribution defined in Section \ref{sec:data},
\begin{align}
&f_{\bf w}\left(U_{\text{stacked}}\right)_{0:T,j}=\sum\limits_{i=1}^{n}a_i\boldsymbol{W}^{V *} \cdot\left(\frac{1}{N} \sum_{p=1}^N \mathbf{1}_{\left(U_i\right)_{0 p}=1}, \cdots, \frac{1}{N} \sum_{p=1}^N \mathbf{1}_{\left(U_i\right)_{Tp}=1}\right)\nonumber\\
&\quad+a_{n+1}\boldsymbol{W}^{V *}\left(\frac{1}{N}\sum\limits_{p=1}^{N}\boldsymbol{1}_{\left({\widetilde{U}}_{\text q}\right)_{0p}=1},\cdots,\frac{1}{N}\sum\limits_{p=1}^{N}\boldsymbol{1}_{\left({\widetilde{U}}_{\text q}\right)_{T+K+1,p}=1}\right)^{\top}\nonumber\\
&=\sum\limits_{i=1}^{n}a_i\cdot\boldsymbol{W}^{V *}\cdot\left(0,\frac{1-p_m}{T}+p_m,\frac{1-p_m}{T},\cdots,\frac{1-p_m}{T}\right)^{\top}+a_{n+1}\cdot\boldsymbol{W}^{V *}\cdot\left(p_m, \frac{1-p_m}{T}, \cdots, \frac{1-p_m}{T}\right)^{\top}\nonumber\\
&=\sum\limits_{j=1}^{n}a_j\cdot \boldsymbol{W}^{V *}\cdot\left(0,\frac{1-p_m}{T}+p_m,\frac{1-p_m}{T},\cdots,\frac{1-p_m}{T}\right)^{\top}+a_{n+1}\cdot\boldsymbol{W}^{V *}\cdot\left(p_m, \frac{1-p_m}{T}, \cdots, \frac{1-p_m}{T}\right)^{\top}\label{eq:prompttopic}
\end{align}
Now consider each entry in \eqref{eq:prompttopic}.

For simplicity, we assume $t^{*}=1$. Then with the optimal $\boldsymbol{W}^{V*}$ in \eqref{eq:optimalt} and \eqref{eq:optimalk}, we can derive:
\begin{align}
&\quad\sum_{i=1}^n a_i \cdot \boldsymbol{W}^{V *} \cdot\left(0, \frac{1-p_m}{T}+p_m, \frac{1-p_m}{T}, \cdots, \frac{1-p_m}{T}\right)^{\top}_1\nonumber\\
&=\sum_{i=1}^n a_i\cdot \left(\frac{-u^{*}\left(1-p_m\right)}{p_m}\times 0+(u^{*}+\frac{1}{1-p_m})\times\left(\frac{1-p_m}{T}+p_m\right)+(T-1)\cdot u^{*}\times \frac{1-p_m}{T}\right)\nonumber\\
&=\sum_{i=1}^n a_i\cdot \left(u^{*}+\frac{p_m}{1-p_m}+\frac{1}{T}\right)\label{eq:topic1}
\end{align}
For $2\leqslant l \leqslant T$,
\begin{align}
&\quad\sum_{i=1}^n a_i \cdot \boldsymbol{W}^{V *} \cdot\left(0, \frac{1-p_m}{T}+p_m, \frac{1-p_m}{T}, \cdots, \frac{1-p_m}{T}\right)^{\top}_l\nonumber\\
&=\sum_{i=1}^n a_i\cdot\left(\frac{-u^*\left(1-p_m\right)}{p_m} \times 0+\left(u^*+\frac{1}{1-p_m}\right)\times\frac{1-p_m}{T}+u^{*}\times\left(\frac{1-p_m}{T}+p_m\right)+(T-2)\cdot u^{*}\times\frac{1-p_m}{T}\right)\nonumber\\
&=\sum_{i=1}^n a_i\cdot \left(u^{*}\times \frac{1-p_m}{T}+\frac{1}{T}+u^* \times\frac{1-p_m}{T}+u^{*}p_m+(T-2) \cdot u^* \times \frac{1-p_m}{T}\right)\nonumber\\
&=\sum_{i=1}^n a_i\cdot\left(\frac{1}{T}+u^{*}\right)\label{eq:topic2_T}
\end{align}
Plug \eqref{eq:topic1} and \eqref{eq:topic2_T} to \eqref{eq:prompttopic}, we can derive:
\begin{align}
&\quad f_{\bf w}\left( U_{\text{stacked}}\right)_{0:T,j}\nonumber\\
&=\sum_{i=1}^n a_i \cdot \boldsymbol{W}^{V_*} \cdot\left(0, \frac{1-p_m}{T}+p_m, \frac{1-p_m}{T}, \cdots, \frac{1-p_m}{T}\right)^{\top}+a_{n+1} \cdot \boldsymbol{W}^{V_*} \cdot\left(p_m, \frac{1-p_m}{T}, \cdots, \frac{1-p_m}{T}\right)^{\top}\nonumber\\
&=\sum_{i=1}^n a_i \cdot \left(0,u^*+\frac{p_m}{1-p_m}+\frac{1}{T},\frac{1}{{T}}+u^*,\cdots,\frac{1}{T}+u^*\right)^{\top}+a_{n+1}\left(0,\frac{1}{T},\cdots,\frac{1}{T}\right)^{\top}\nonumber\\
&=\left(0,\sum_{i=1}^n a_i\cdot u^*+\frac{p_m}{1-p_m}+\frac{1}{T},\sum_{i=1}^n a_i\cdot u^{*}+\frac{1}{T},\cdots,\sum_{i=1}^n a_i\cdot u^{*}+\frac{1}{T}\right)^{\top}\label{eq:topicpredprompt}
\end{align}
From the expression in \eqref{eq:topicpredprompt}, we know the largest entry in the vector $f_{\bf w}\left( U_{\text{stacked}}\right)_{0:T,j}$ corresponds to $k^*_{\text q}=1$.

Now we consider the class labels. 
\begin{align*}
&f_{\bf w}\left(U_{\text{stacked}}\right)_{T+1:T+K+1,j}=\sum\limits_{i=1}^{n}a_i\boldsymbol{W}^{V *} \cdot\left(\frac{1}{N} \sum_{p=1}^N (U_i)_{T+1,p}, \cdots, \frac{1}{N} \sum_{p=1}^N (U_i)_{T+K+1,p}\right)\nonumber\\
&\quad+a_{n+1}\boldsymbol{W}^{V *}\left(\frac{1}{N}\sum\limits_{p=1}^{N}{\left({\widetilde{U}}_{\text q}\right)_{T+1,p}},\cdots,\frac{1}{N}\sum\limits_{p=1}^{N}\left({\widetilde{U}}_{\text q}\right)_{T+K+1,p}\right)^{\top}\nonumber\\
&=\sum\limits_{i=1}^{n}a_i\boldsymbol{W}^{V*}\cdot\left(0,Q\cdot\boldsymbol{1}_{k^{*}(U_i)=1}+\frac{1-Q}{K-1}\cdot\boldsymbol{1}_{k^{*}(U_i)\neq 1},\frac{1-Q}{K-1}, \cdots,\frac{1-Q}{K-1}\cdot\boldsymbol{1}_{k^{*}(U_i)\neq 1},\frac{1-Q}{K-1}\right)\\
&\quad +a_{n+1}\cdot\left(p_m,Q(1-p_m), \frac{1-Q}{K-1}(1-p_m),\cdots,\frac{1-Q}{K-1}(1-p_m)\right)
\end{align*}
So $\forall l=T+2,\cdots,T+K+1$, denote $\Lambda_k:=\left\{i: k^*\left(U_i\right)=1\right\}$, where $k^*(\cdot)$ is the class of first token in a sequence.
\begin{align*}
&Q\cdot\sum\limits_{i\in\Lambda_1}a_i+\frac{1-Q}{K-1}\sum\limits_{i\notin \Lambda_1}a_i\cdot+Q(1-p_m)\cdot a_{n+1}:=b_{T+2}
\end{align*}
\begin{align*}
&Q\cdot\sum\limits_{i\in\Lambda_l}a_i+\frac{1-Q}{K-1}\sum\limits_{i\notin \Lambda_l}a_i\cdot+\frac{1-Q}{K-1}(1-p_m)\cdot a_{n+1}:=b_{l}  
\end{align*}
In this case, we have
\begin{align*}
&f_{\bf w}(U_{\text{stacked}})_{T+k+1, j}=-\frac{q^*(1-p_m)}{p_m}\cdot b_{T+1}+\left(q^{*}+\frac{1}{1-p_m}\right)\cdot b_k+q^{*}\cdot\sum\limits_{r\neq k}b_{r}.
\end{align*}
Since $q^*+\frac{1}{1-p_m}>q^*$, suppose  $b_{T+2}>b_k,k\neq T+2$, then we have $$\underset{k}\arg\max f_{\bf w}( U_{\text{stacked}})_{T+k+1,j}=1$$.
In fact, there always exists $a_i$ such that $b_{T+2}>b_k$ holds, if we choose large enough $a_{n+1}$.
\end{proof}
\begin{lemma}\label{lemma:S}
Let $S$ denote the set of solution in \eqref{eq:optimalt} and \eqref{eq:optimalk} that $L(\boldsymbol{W}^V)$ We have the following conclusion:
\begin{align*}
 \forall \boldsymbol{W}^{V *} \in\lim _{\lambda \rightarrow 0} \operatorname{argmin} L_{\ell_2}\left(\boldsymbol{W}^V\right),\text{there is }\boldsymbol{W}^{V *} \in S   
\end{align*}
\end{lemma}
The above lemma directly comes from \citep{li2023transformers}.
\section{Experiment Details for Synthetic Data}\label{sec:synthetic}
\subsection{Data Generation}
The vocabulary consists of $100$ words, with $T=10$ topics, denoted as topic `0' to `9'. Each topic has $K=10$ words, denoted as class `0' to `9'. The number of training sequences and test sequences are $10,000$, respectively. We also generate another $10,000$ samples to be the context sample for each test sample. The sequence length is set to be a random number between $100$ and $150$. For each training sequence, we randomly draw a key topic $t$, with probability $0.55$, the word is drawn from topic $t$. The class generation follows from our setting in Section \ref{sec:data} with the key class probability $Q=0.91$, and other $9$ classes occur with probability $\frac{1-Q}{K-1}=0.01$. Randomly mask out $15\%$ of the words in each training sequence. For the test sequences, we set $L_1=0.7*N$ and $L_2=0.3*N$, the distribution follow Section \ref{sec:data} of $\widetilde{D}_{\text q}$. The context samples follow ${D}_{\text q}$ with $L_1=0.7*N$ and $L_2=0.3*N$. Eech test and context sample pair share the same key topic $t^{*}$ in Section \ref{sec:data}. Build the prompt by stacking the test and context sequence pair.
\subsection{Training Hyper parameters}
Our experiment set up follows from \citep{li2023transformers}. We use a one hidden-layer Bert with one attention head and hidden dimension $D=104$. Use Adam optimizer with learning rate $0.01$. The batch size is set to be $40$.
\subsection{Compare Uniform Attention and Non-Uniform Attention}

\begin{figure}[ht]
	\begin{center}
        \includegraphics[width=0.45\linewidth]{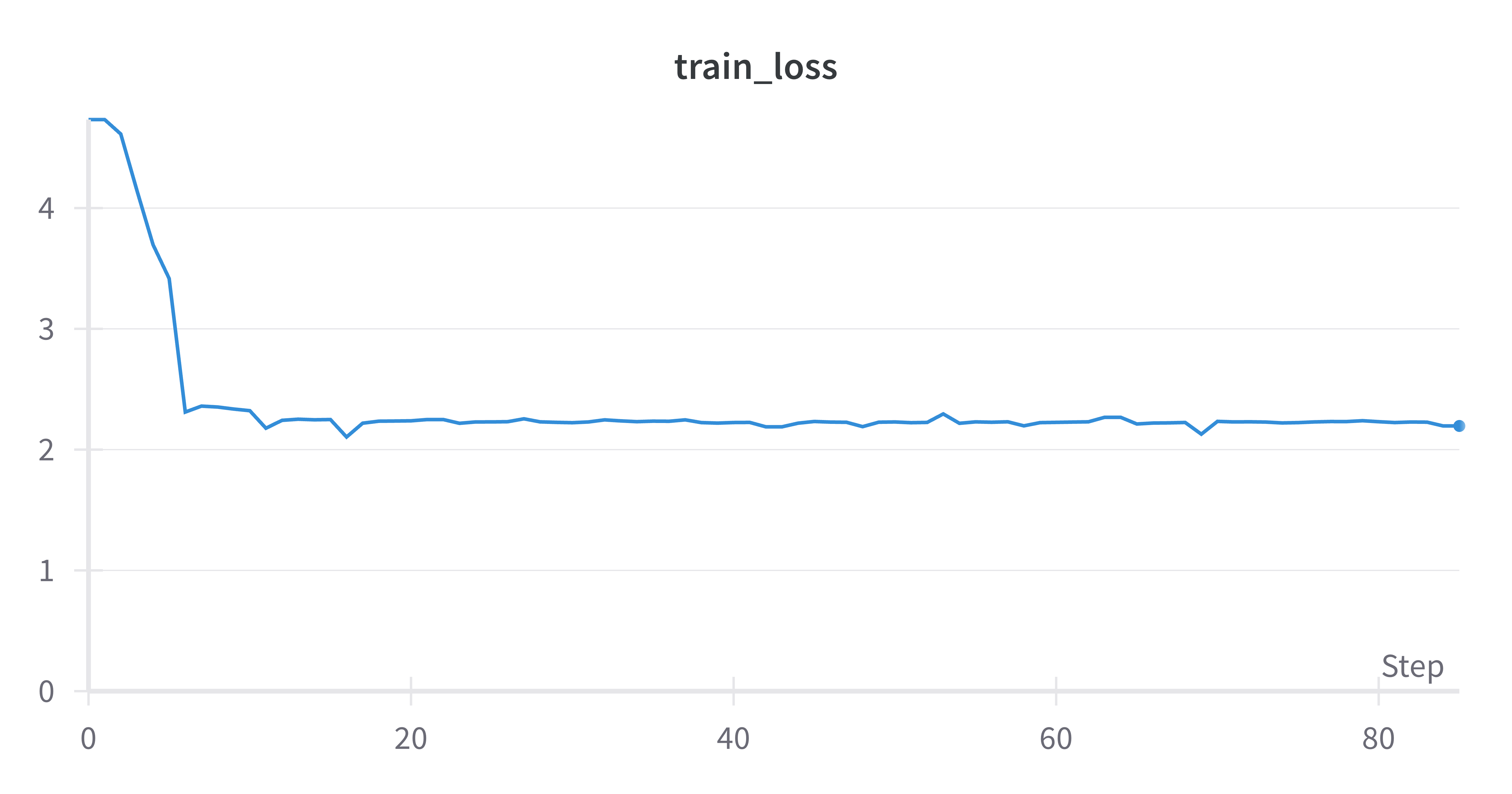}
        \hspace{-2mm}
        \includegraphics[width=0.45\linewidth]{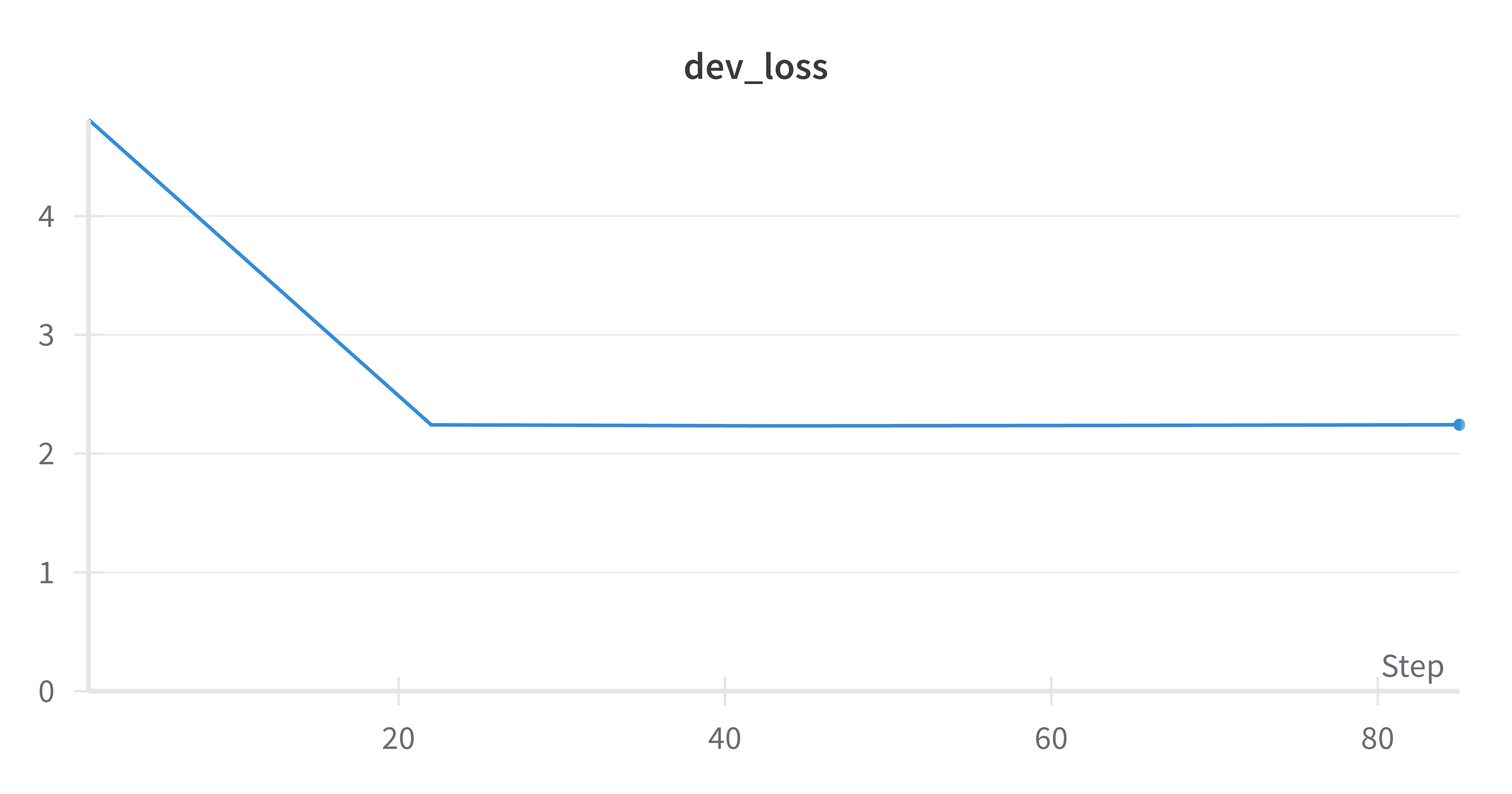}
	\end{center}
	\caption{The training loss and dev loss when the network is trained with uniform attention, i.e, $\boldsymbol{W}^Q=\boldsymbol{W}^K=0$.}
	\label{fig:text_pathfinder}
\end{figure}
\begin{figure}[ht]
	\begin{center}
        \includegraphics[width=0.45\linewidth]{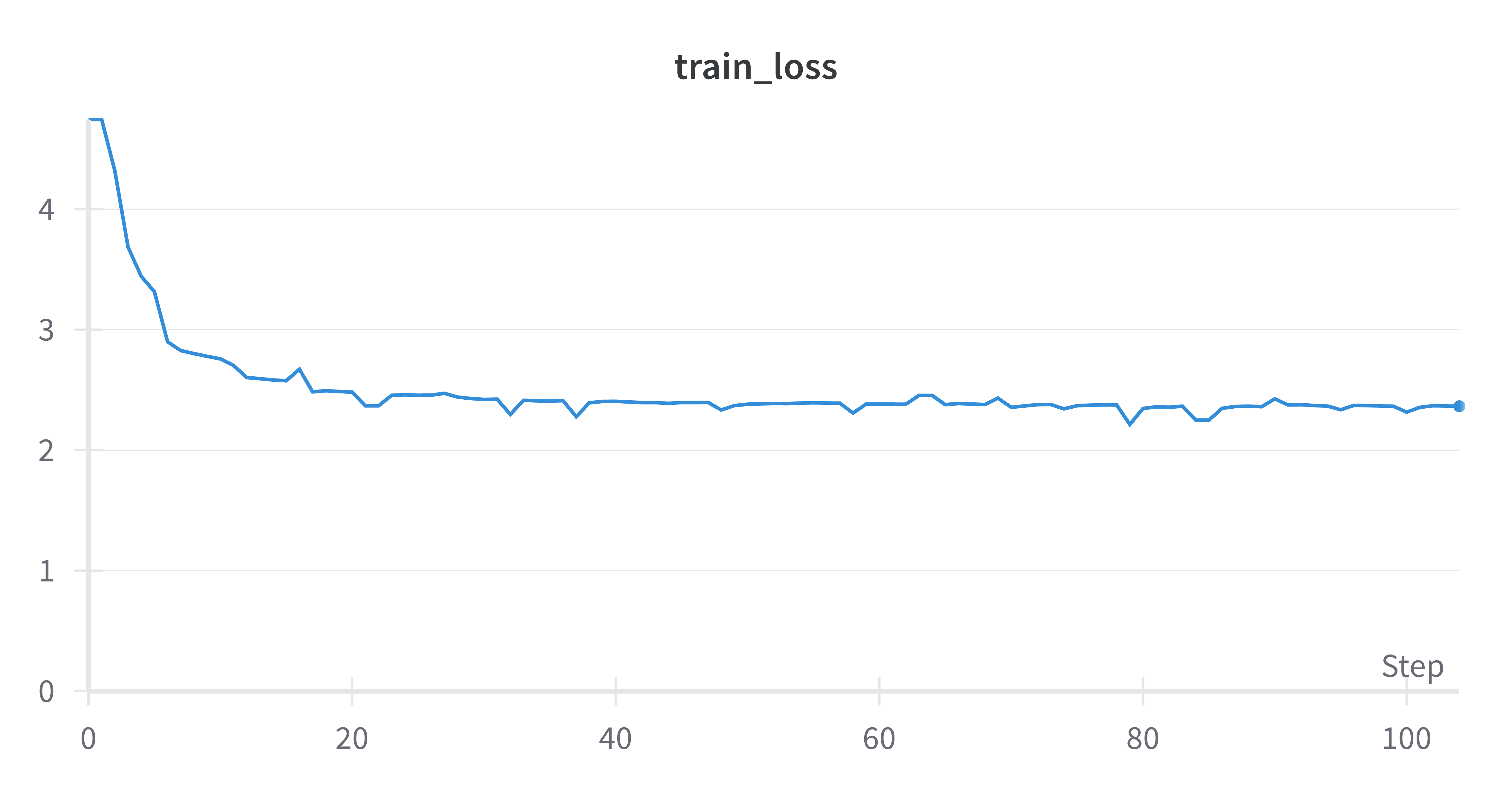}
        \hspace{-2mm}
        \includegraphics[width=0.45\linewidth]{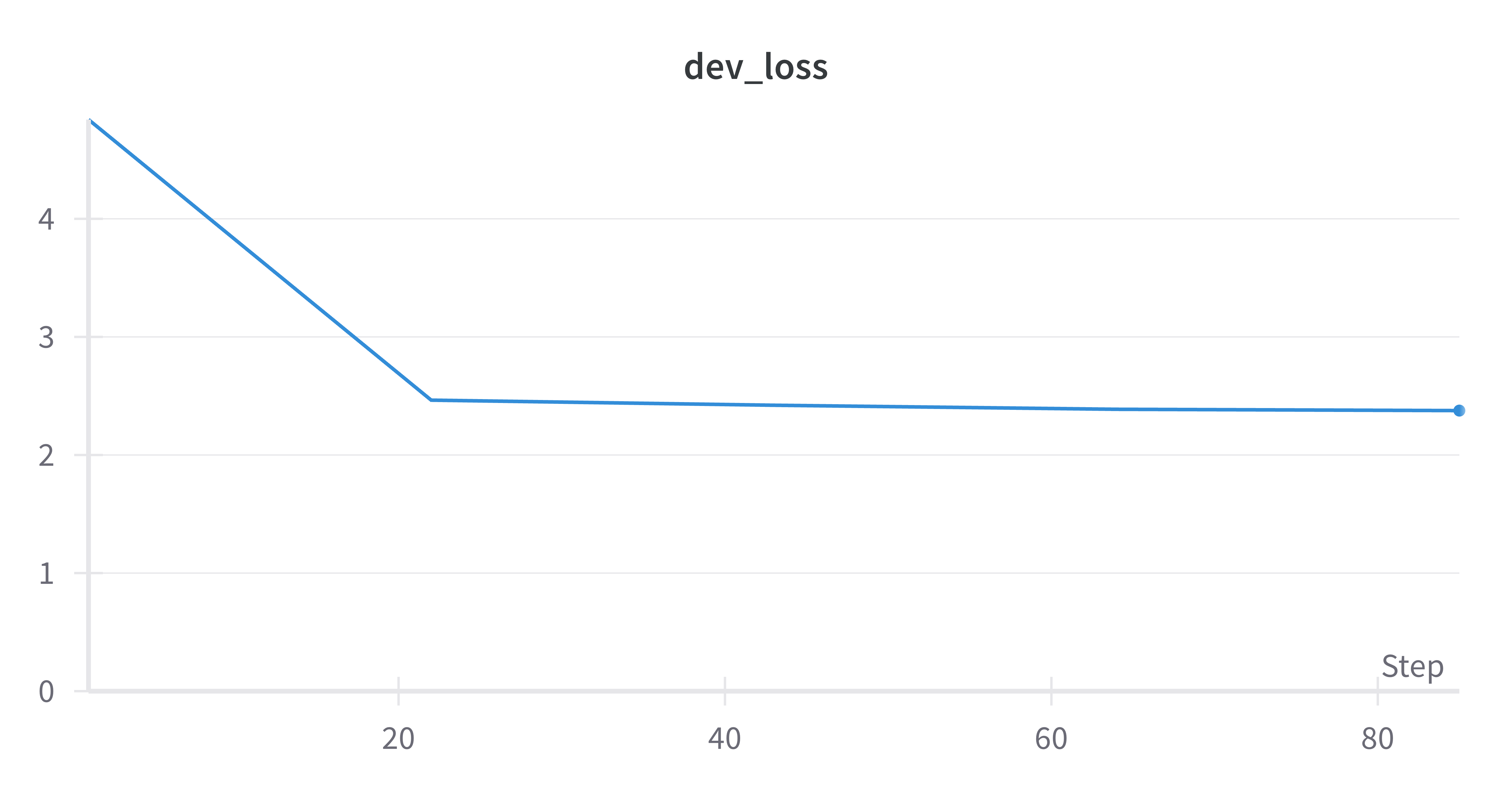}
	\end{center}
	\caption{The training loss and dev loss when the network is trained with  $\boldsymbol{W}^Q,\boldsymbol{W}^K$ updated.}
	\label{fig:text_pathfinder2}
\end{figure}
We conclude from Fig. \ref{fig:text_pathfinder} and \ref{fig:text_pathfinder2} that, freezing the attention to be uniform in the synthetic experiment setting does not influence the training or test performance. Thus, we justify that it is acceptable when we freeze the attention head in our analysis.
\section{Experiment Details for GPT-2 Models}\label{sec:gpt}
\textbf{Details of concept token training:} To train for the concept token for each fine-tuning task/dataset, we add $N = 5$ concept tokens into the input layer of the GPT-2 model, essentially extending the total number of tokens by $N=5$. 
Then we fine-tune on each task/dataset, with only the embedding layer of the GPT-2 model trainable 
to minimize the prediction error. This will result in $N=5$ new token embeddings that could represent the concept of the task/dataset. The detailed process is illustrated by the ICL framework in \citet{min-etal-2022-metaicl}. The trained GPT-2 model is discarded in the further experiment steps, while the concept token and its corresponding embedding vector (called representation vector in the main text) are kept for further experiment steps\\
\textbf{Hyperparameters:} We use seven different concept datasets to fine-tune the modified GPT-2 large embedding layer. The model is saved after 10,000 training steps. All seven concept tokens are passed through the embedding layer, getting 7 * 5 * 1280 matrix, {number of task * tokens for each task * embedding length}, to represent seven concepts. We use the t-SNE \citep{van2008visualizing} method to reduce the dimension to 7 * 5 * 2 with perplexity $p = 5$, and calculate the distance of each concept. After selecting concepts that are similar or dissimilar to the {target} task based on distance, we use the datasets corresponding to these two sets of concepts to fine-tune the entire model. In {test}, we use $k = 16$ number of demonstrations for our experiment. We organize the {target} datasets into batches, with each batch comprising 8 samples grouped together, and run the {test} with seed 100. 
The code we use for in-context inference is based on \citep{min-etal-2022-metaicl}.\\
\textbf{Concept Token Visualization: } 
After we use t-SNE to reduce the dimension to 5 * 2 for each concept, we use the first column as the x-coordinate and the second column as y-coordinate to print five points as in Figure \ref{fig:concept_distance}. The five concepts corresponding to the same task are printed in the same color.
\begin{table}[h]
\centering
\resizebox{\columnwidth}{!}{%
    \begin{tabular}{|l|l|l|l|l|l|l|l|l|l|}
    \hline
        ~ & hate\_speech\_offensive & hatexplain & tweet\_eval-hate & tweet\_eval-offensive & ag\_news & glue-sst2 & dream \\ \hline
        hate\_speech\_offensive & 0.000 & 117.051 & 129.666 & 143.543 & 218.394 & 133.639 & 284.768  \\ \hline
        hatexplain & 117.051 & 0.000 & 97.176 & 94.166 & 148.673 & 163.069 & 208.924   \\ \hline
        tweet\_eval-hate & 129.666 & 97.176 & 0.000 & 72.043 & 193.365 & 174.350 & 249.684  \\ \hline
        tweet\_eval-offensive & 143.543 & 94.166 & 72.043 & 0.000 & 169.720 & 160.353 & 256.166 \\ \hline
        ag\_news & 218.394 & 148.673 & 193.365 & 169.720 & 0.000 & 171.542 & 138.502  \\ \hline
        glue-sst2 & 133.639 & 163.069 & 174.350 & 160.353 & 171.542 & 0.000 & 236.988  \\ \hline
        dream & 284.768 & 208.924 & 249.684 & 256.166 & 138.502 & 236.988 & 0.000  \\ \hline
    \end{tabular}%
    }

\caption{Concept distance of the seven tasks. We calculate the distances between two tasks by computing the distance between the five tokens of the two group of concepts (note that each task has 5 concept token embeddings), and select the minimum value as our final calculated distance.} 
\label{tab:concept_distance}
\end{table}

\begin{table}[h]
\centering
\resizebox{\columnwidth}{!}{%
    \begin{tabular}{|l|l|l|l|l|l|l|l|l|l|}
    \hline
        ~ & ag\_news & yelp\_polarity & imdb & quoref & glue-cola & rotten\_tomatoes & wiki\_qa & tweet\_eval-offensive \\ \hline
        ag\_news & 0.000 & 4.095 & 4.200 & 3.927 & 6.786 & 4.272 & 4.193 & 2.050  \\ \hline
        yelp\_polarity & 4.095 & 0.000 & 2.096 & 5.856 & 5.302 & 1.997 & 5.470 & 4.973   \\ \hline
        imdb & 4.200 & 2.096 & 0.000 & 4.818 & 4.057 & 1.848 & 4.840 & 5.239 \\ \hline
        quoref & 3.927 & 5.856 & 4.818 & 0.000 & 4.469 & 4.708 & 2.027 & 5.763 \\ \hline
        glue-cola & 6.786 & 5.302 & 4.057 & 4.469 & 0.000 & 3.742 & 4.704 & 8.447  \\ \hline
        rotten\_tomatoes & 4.272 & 1.997 & 1.848 & 4.708 & 3.742 & 0.000 & 4.591 & 5.465  \\ \hline
        wiki\_qa & 4.193 & 5.470 & 4.840 & 2.027 & 4.704 & 4.591 & 0.000 & 5.931 \\ \hline
        tweet\_eval-offensive & 2.050 & 4.973 & 5.239 & 5.763 & 8.447 & 5.465 & 5.931 & 0 \\ \hline
    \end{tabular}%
    }

\caption{Concept distance of the eight other tasks using the same method as Table \ref{tab:concept_distance}.} 
\label{tab:concept_distance2}
\end{table}
\section{Comparison of Predictions from Different Prompts }\label{sec:pred}

 \noindent\textbf{(a) Prediction from \eqref{eq:prompt}:} Set $\mathbf{Z}=\mathbf{Z}_{\text{stacked}}$ in \eqref{eq:transformer}. In this case, $L=D,G=2n+2$.
\begin{align}
\widehat{y}_{\text {q}}&=f_{\bf w}(\mathbf{Z})_{1:D, 2n+2}=\left(\boldsymbol{W}^V X_1,\boldsymbol{W}^V y_1,\boldsymbol{W}^V X_2,\boldsymbol{W}^V y_2, \cdots, W^V X_{\text q}, \mathbf{0}\right)\cdot \sigma\left(\begin{array}{c}
X_1^{\top} \cdot B \cdot \mathbf{0} \\
y_1^{\top} \cdot B \cdot \mathbf{0} \\
\cdots \\
X_{\text q}^{\top} \cdot B \cdot \mathbf{0}\\
\mathbf{0}^{\top}\cdot B\cdot \mathbf{0}
\end{array}\right)\nonumber\\
&=\frac{1}{2n+2}\sum\limits_{i=1}^{n}W^V(X_i+y_i)+\frac{1}{2n+2}W^V{X}_{\text {q}}^{\top}\label{eq:truepred:1}
\end{align}
\noindent\textbf{(b) Prediction from \eqref{eq:fakeprompt}:} Set $\mathbf{Z}=\mathbf{Z}_{\text{stacked}}$ in \eqref{eq:transformer}. In this case, $L=2D,G=n+1$.
\begin{align}
\widehat{y}_{\text {q}}=f_{\bf w}(\boldsymbol{Z})_{D+1:2D,n+1}=\left(\boldsymbol{W}^{V'} \boldsymbol{Z}\right) \cdot\mathcal{M}\cdot\sigma\left(\frac{\left(\boldsymbol{W}^{K'} \boldsymbol{Z}\right)^{\top}\left(\boldsymbol{W}^{Q'} \boldsymbol{Z}\right)}{\sqrt{2D}}\right),\;\mathcal{M}:=\left(\begin{array}{cc}
I_n & 0 \\
0 & 0
\end{array}\right) \in \mathbb{R}^{(n+1) \times(n+1)}.\nonumber
\end{align}
More specifically, in \citep{zhang2023trained}, the weight has following structure:
\begin{align}
\boldsymbol{W}^{V'}=\left[\begin{array}{cc}
\mathbf{0}_{D \times D} & \mathbf{0}_{D \times D} \\
\mathbf{0}_{D \times D} & I_{D}
\end{array}\right]\in\mathbb{R}^{2D\times 2D}, \quad {\boldsymbol{W}^{K'}}^{\top}{\boldsymbol{W}^{Q'}}=\left[\begin{array}{cc}
B' & \mathbf{0}_{D \times D} \\
\mathbf{0}_{D \times D} & \mathbf{0}_{D \times D}
\end{array}\right] \quad \text { where } B' \in \mathbb{R}^{D \times D} \text {. }
\end{align}
Thus, we can derive 
\begin{align}
\widehat{y}_{\text q}=\left(y_1,y_2,\cdots,y_n,\mathbf{0}\right)\cdot\sigma\left(\begin{array}{ll}
X_1^{\top}\cdot B'\cdot X_{\text q}\\
X_2^{\top}\cdot B'\cdot X_{\text q}\\
\quad\quad\cdots\\
X_n^{\top}\cdot B'\cdot X_{\text q}
\end{array}\right)=\sum\limits_{i=1}^{n}\sigma_iy_i,\label{eq:fakepred:2}
\end{align}
where $\sum\limits_{i=1}^{n}\sigma_i=1$, and each $\sigma_i=\frac{\exp \left(X_i^{\top} \cdot B^{\prime} \cdot X_{\text q}\right)}{\sum\limits_{l=1}^n \exp \left(X_l^{\top} \cdot B^{\prime} \cdot X_{\text {q}}\right)}$ .
\section{Limitation and Future Work}\label{sec:limit}
In our work, we requires each class within a topic only includes one word. We make this assumption to simplify our analysis. However, in more realistic setting, one class  can contain multiple words, which requires a more complicated modeling. Further, we assume uniform attention in our network, which can be extended to more realistic setting where $W^Q,W^K$ are updated. These extension will left to future work.
\newpage

\end{document}